\documentclass{article}

\usepackage[preprint]{neurips_2025}

\usepackage[utf8]{inputenc} 
\usepackage[T1]{fontenc}    
\usepackage{hyperref}       
\usepackage{url}            
\usepackage{booktabs}       
\usepackage{amsfonts}       
\usepackage{nicefrac}       
\usepackage{microtype}      
\usepackage{xcolor}         
\usepackage{comment}
\usepackage{graphicx}
\usepackage{adjustbox}
\usepackage{multirow,multicol}
\usepackage{amsmath}
\usepackage{colortbl}
\definecolor{comfortgreen}{rgb}{0.75, 0.85, 0.75}
\definecolor{mistyrose}{rgb}{1.0, 0.89, 0.88}
\definecolor{bluebar}{rgb}{0.337, 0.706, 0.914}
\definecolor{greenbar}{rgb}{0.8, 0.910, 0.271}

\title{How Do Vision-Language Models Process Conflicting Information Across Modalities?}

\author{Tianze Hua\thanks{Equal contribution.} \\
  Brown University \\
  \texttt{tianze\_hua@brown.edu} \\\And
  Tian Yun\footnotemark[1] \\
  Brown University \\
  \texttt{tian\_yun@brown.edu} \\\And
  Ellie Pavlick \\
  Brown University \\
  \texttt{ellie\_pavlick@brown.edu}
}

\begin{document}

\maketitle

\begin{abstract}
    AI models are increasingly required to be multimodal, integrating disparate input streams into a coherent state representation on which subsequent behaviors and actions can be based. This paper seeks to understand how such models behave when input streams present conflicting information. Focusing specifically on vision-language models, we provide inconsistent inputs (e.g., an image of a dog paired with the caption ``A photo of a cat'') and ask the model to report the information present in one of the specific modalities (e.g., ``What does the caption say / What is in the image?''). We find that models often favor one modality over the other, e.g., reporting the image regardless of what the caption says, but that different models differ in which modality they favor. We find evidence that the behaviorally preferred modality is evident in the internal representational structure of the model, and that specific attention heads can restructure the representations to favor one modality over the other. Moreover, we find modality-agnostic ``router heads'' which appear to promote answers about the modality requested in the instruction, and which can be manipulated or transferred in order to improve performance across datasets and modalities. Together, the work provides essential steps towards identifying and controlling if and how models detect and resolve conflicting signals within complex multimodal environments. All code and resources are available at: \url{https://github.com/ethahtz/vlm_conflicting_info_processing}

\end{abstract}

\section{Introduction}

With the recent and substantial advances in AI, there is increasing interest in building richly multimodal AI systems capable of carrying out complex tasks involving inputs beyond just text, from online purchasing \citep{openai_operator_2025} to biomedical monitoring \citep{acosta2022multimodal} to mental health support \citep{heinz2024evaluating}.
Such systems will need to integrate a variety of modalities into a coherent internal state representation in order to make intelligent decisions. In realistic environments, these different input streams are likely to present conflicting information for a number of reasons -- faulty or noisy sensors \citep{liao2025benchmarkingmultimodalsemanticsegmentation}, human errors in data reporting, and even malicious or adversarial attacks \citep{10350690, cui2024robustness, zhu2025callingspadeheartgaslighting}. Thus, it is important to understand how multimodal models respond when given conflicting information across modalities.

To study this question, we focus on vision-language models (VLMs), which have achieved impressive performance across a wide range of tasks, from visual question answering to image captioning and grounded instruction following \citep{10.5555/3600270.3601993,liu2023improvedllava,dai2023instructblip,openai2024gpt4technicalreport,Qwen2.5-VL}. We construct conflicting image-caption pairs from existing image recognition datasets \citep{krizhevsky2009learning, deng2009imagenet, pascal, wah_branson_welinder_perona_belongie_2011} and prompt models to report the information of the requested modality based on the given inconsistent image and caption inputs. Using a range of analysis techniques, we investigate how models respond to these inputs and begin to uncover the mechanisms responsible for the observed behaviors. Specifically, we make the following contributions:

\begin{enumerate}
    \item We evaluate a variety of VLMs and demonstrate that all models show performance degredation when presented with inconsistent image and caption inputs. Contrary to related concurrent work \citep{deng2025wordsvisionvisionlanguagemodels}, our results show that models vary in their biases, in some settings trusting the caption over the image and others settings doing the opposite (Section \ref{sec:behavioral_results}).
    \item Using a variety of probing and clustering techniques, we show that the VLM internal representations encode significant information about each modality independently, but that they are not equally well encoded. Indeed, we find that a model's bias toward image over caption can be directly observed in the representational structure of the mid- to late- layers of the network (Section \ref{sec:representational_analysis}).
    \item Finally, we localize individual attention heads that are responsible for dynamically restructuring these internal representations and thus influencing the model's observed biases. We identify two distinct head types -- modality-agnostic \textit{router heads} and modality-specific \textit{promotion heads} -- which can be manipulated or transferred across datasets and modalities in order to influence model performance (Section \ref{sec:head_attribution}). 
\end{enumerate}

\section{Performance Degradation of VLMs on Conflicting Inputs} \label{sec:behavioral_results}

\begin{figure}[t]
    \centering
    \includegraphics[width=1.0\linewidth]{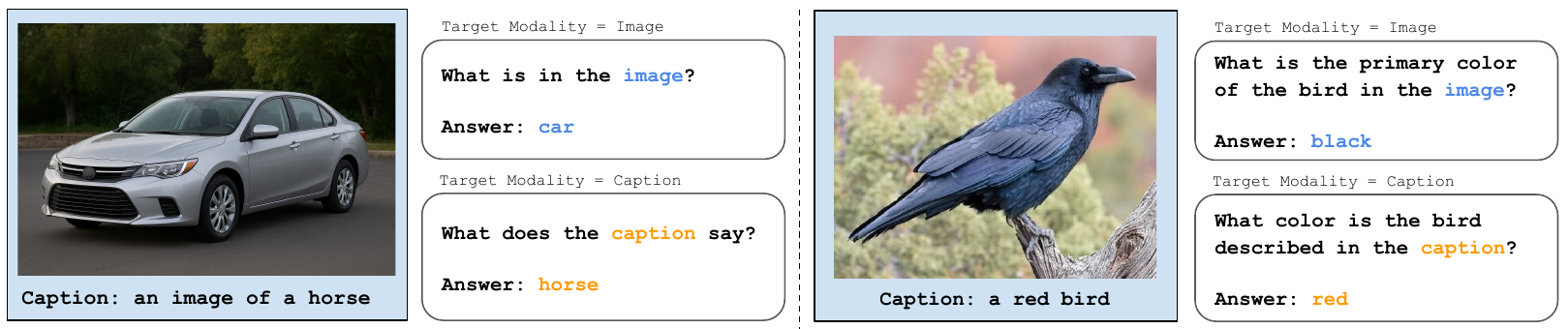}
    \caption{Examples of inconsistent image and caption pairs. Left: image and caption disagrees on the main object of the image; right: image and caption disagrees on an attribute of the main object of the image. Either case, model needs to report image or caption information based on the target modality.}
    \label{fig:task_example}
\end{figure}

\paragraph{Experimental Design}
To study how VLMs process information when presented with conflicting inputs across modalities, we introduce the following task setup: given a pair of inconsistent image and caption, the model is prompted to report information from a designated \textit{target modality}, where the corresponding content in the non-target modality is \textit{misleading} (Figure~\ref{fig:task_example}). 

By analyzing model behavior on this task, we aim to assess whether such input inconsistency poses challenges for unimodal information reporting, and whether models exhibit inherent preferences for one modality over another.

\paragraph{Datasets and Construction}
To construct conflicting image and caption pairs, we leverage existing image classification and object/attribute detection datasets and adversarially sample image classes/attributes that are inconsistent with the content of each image. Specifically, we use four standard image classification datasets: CIFAR-10, CIFAR-100 \citep{krizhevsky2009learning}, ImagetNet100 \citep{deng2009imagenet}, and Pascal VOC \citep{pascal}. In addition, we construct a subset of the CUB dataset (CUB-Color) \citep{wah_branson_welinder_perona_belongie_2011} by filtering for bird images associated with a single primary color attribute, ensuring each image corresponds to exactly one dominant color label\footnote{There exists differences in how realistic a conflict between an image and caption pair is. In CUB-Color, the primary color of birds is more subtle; while in the other four datasets, a wrong caption is more artificial (e.g., ``an image of a horse'' paired with an image of a car).}.

Given a consistent pair of image and class label $(\texttt{IMAGE}_i,c_i)$ and a set of all possible classes $C$, we randomly sample a false class label $\tilde c_i \sim \mathrm{Uniform}\bigl(C\setminus\{c_i\}\bigr)$ and pair it with $\texttt{IMAGE}_i$ to form an inconsistent pair of image and class label $(\texttt{IMAGE}_i,\tilde c_i)$. After obtaining these inconsistent pairs, we implement a prompt generator to generate the textual prompt for the models given their instruction templates, as well as the target modality. 

For the rest of the paper, when we refer to a dataset, we mean the dataset with inconsistent image-caption pairs, and not the original dataset. For more details on prompt templates, see Appendix~\ref{app:prompt_templates}.

\paragraph{Models} We run behavioral evaluation of the task on the following four 7b-parameter multimodal large language models: LLaVA-1.5 \citep{liu2023improvedllava}, InstructBLIP \citep{dai2023instructblip}, Qwen2.5-VL \citep{Qwen2.5-VL} and LLaVA-OneVision \citep{li2025llavaonevision}.

\paragraph{Unimodal Baselines} In addition to evaluating models on inconsistent inputs, we include the unimodal baselines, where a VLM receives an input from a single modality -- either an image or a caption -- rather than a pair of them. For the image baseline, we use a visual question answering style evaluation, where thr VLM takes in an image and a query prompt (e.g., ``What is in the image?''). For the caption baseline, we use only the language model of the VLM. These baselines indicate how well the model can report from each modality by default, in the absence of cross-modal interference.

\begin{figure}[t]
    \centering
    \includegraphics[width=1\linewidth]{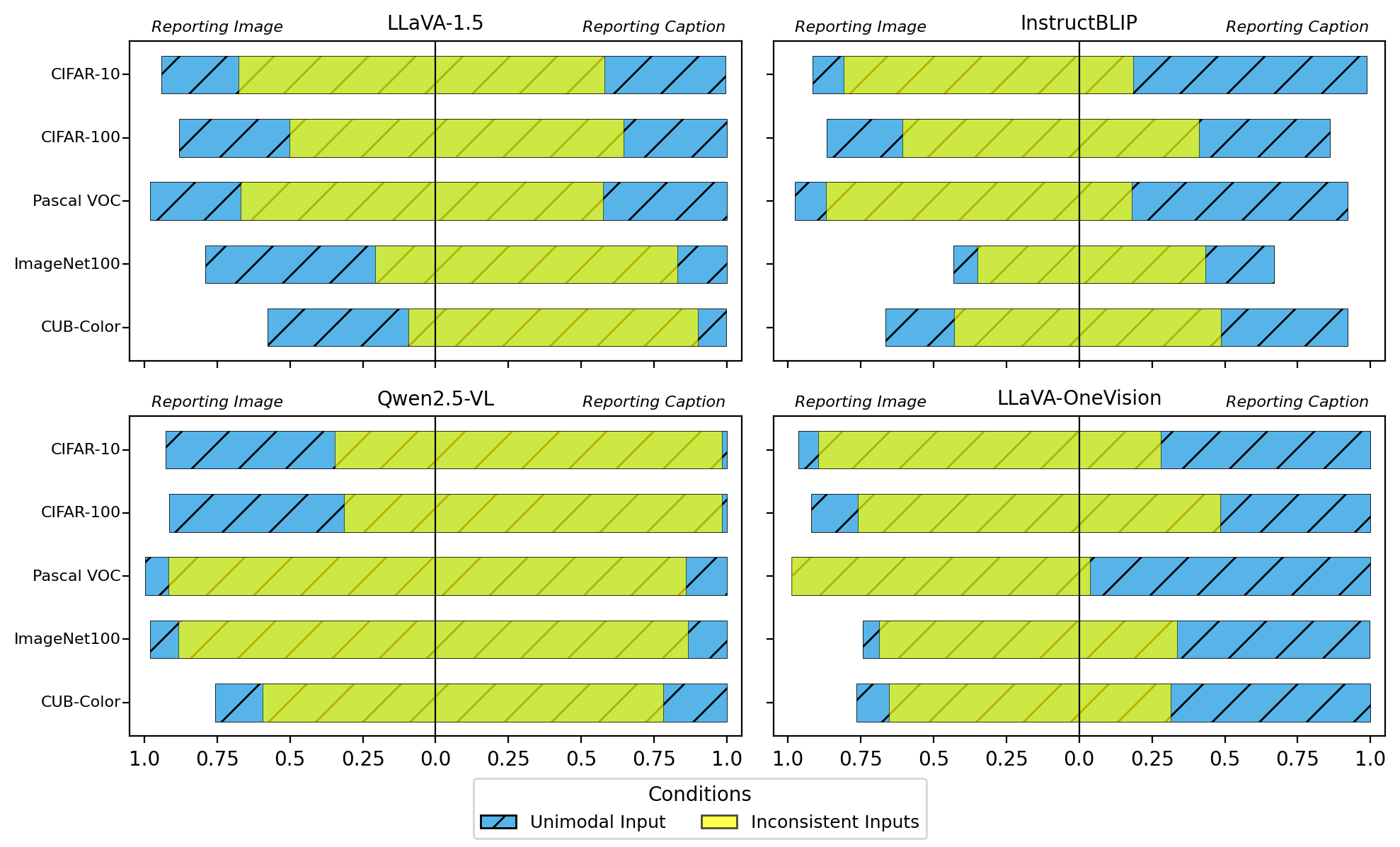}
    \caption{Model performance on reporting target modality information under unimodal inputs and inconsistent inputs. Horizontal bars show accuracy for each model–dataset pair when the model is asked to report either the image (left column in every panel) or the caption (right column). \colorbox{bluebar!70}{Blue bars}: Unimodal baseline (only input from the target modality is provided). \colorbox{greenbar!70}{Green bars}: Inconsistent-input condition. All models illustrate accuracy drop when conflicting information from the other modality is introduced, indicating cross-modal interference between the inputs. 
    }
    \label{fig:behavioral_results}
\end{figure}

\paragraph{Results}

Figure~\ref{fig:behavioral_results} presents model performance on reporting target modality information under unimodal inputs and inconsistent inputs. 
For almost all models across every dataset, the accuracy in reporting target modality information drops when conflicting information is present in the image-caption pairs, indicating the occurrence of cross-modal interference. Different models suffer differently from the conflicting inputs: InstructBLIP and LLaVA-OneVision lose far more accuracy when the task is to report captions, revealing a constant bias towards reporting information in the images. Qwen-VL 2.5 performs poorly on reporting image for the CIFAR datasets, but otherwise maintains balanced accuracy across modalities, suggesting better selectivity to the target modality despite conflicting input. Most of the incorrect predictions are due to predictions of the class label in the non-target modality -- for a breakdown of prediction results, see Appendix~\ref{app:classification_of_behavioral_responses}. In comparison, we found that when image-caption pairs are consistent, it often brings a positive performance improvement compared to the single-modality input baselines, for more details see Figure~\ref{fig:behavior_consistent}.

Our findings offer more nuance to the observations reported in \citet{deng2025wordsvisionvisionlanguagemodels}, where when models are given images paired with inconsistent texts and prompted to report image information, they are often misled by those ``corrupted'' texts. In our figure, LLaVA-1.5 behaves like this on ImageNet100 and CUB-Color, and Qwen2.5-VL also show similar preferences towards texts on CIFAR-10 and CIFAR-100, even when prompted to report image information. Our results further complement their previous results, since we investigate the reciprocal case where the models are prompted to report caption information under misleading visual information. We find that many models (e.g., InstructBLIP and LLaVA-OneVision) show inherent preference towards the visual modality and are unable to faithfully report text information, thus having a ``blind faith'' in image instead. Together, our experiments comprehensively evaluate and reveal models' preference towards specific modalities under different target modalities.

\section{Hypotheses of Performance Degradation under Conflicting Inputs} 
\label{sec:representational_analysis}

To understand \textit{why} the VLMs exhibit such performance drops, we test three hypotheses: (1) model's inability to encode information from both modalities, (2) model's inability to detect the consistency of the inputs, or (3) model's representations fail to reflect the target modality requested in the prompt.

\subsection{Do VLMs Fail to Encode Unimodal Information?} \label{sec:unimodal_probe}

We first hypothesize that the VLM struggles because it has failed to encode the modalities as distinct within its internal representations -- i.e., losing a notion of ``provenance'' for the information in its input. We use linear probes to measure how much unimodal information is encoded in the representations of inconsistent inputs. Specifically, we train two separate probes for each experiment setting, one is trained to recover the image class label and the other to recover the caption class label. 
If probes show high accuracy on a specific modality, it indicates the unimodal information of that modality is readily available given the representations of conflicting inputs.

\begin{figure}[t]
    \centering
    \includegraphics[width=1\linewidth]{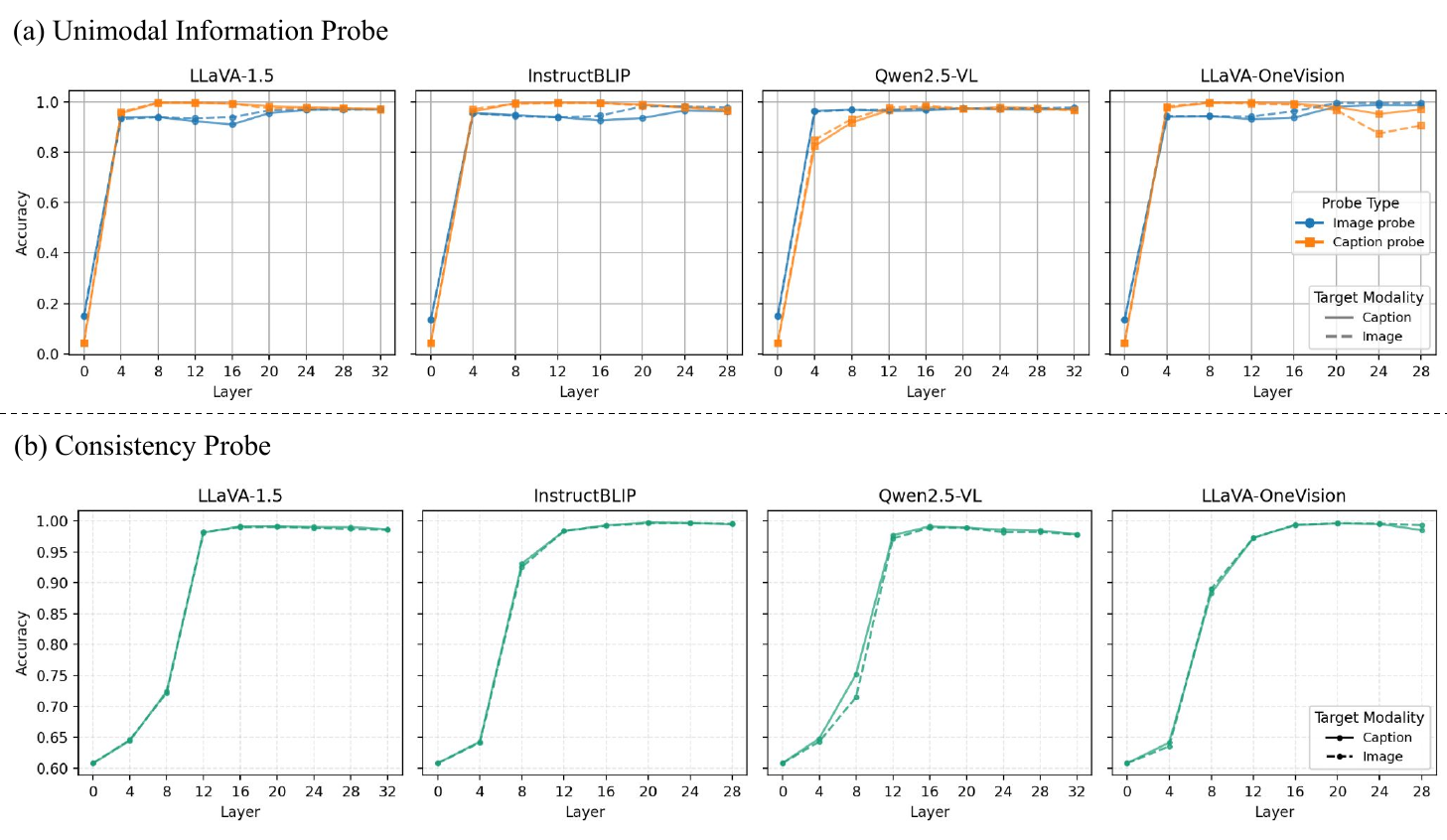}
    \caption{Accuracy of (a) unimodal information probe and (b) consistency probe on Pascal VOC dataset. 
    Probe accuracies remain high on both unimodal information probe and consistency probe on the last layers, indicating that VLMs can sufficiently encode the information of each modality and the information of consistency between two modalities.
    }
    \label{fig:unimodal_probe_accuracy}
\end{figure}

Figure \ref{fig:unimodal_probe_accuracy}(a) shows the unimodal probe accuracy on Pascal VOC dataset. The pattern is consistent across all models: after the early layers, probe accuracy for both image and caption class labels is very high, regardless of the requested target modality. This suggests these representations encode unimodal information for both modalities sufficiently well, and thus requires another explanation for the observed performance drops.

\subsection{Do VLMs Fail to Detect the Inconsistencies between Two Modalities?} 
\label{sec:alignment_probe}

If a VLM is unaware of the consistency of an image-caption pair, then it can assume the current inputs are consistent and report information of the non-target modality when encountering inconsistent inputs.
We use linear probes to measure whether a VLM encodes information about the consistency between modalities in its internal representations of consistent and inconsistent inputs. Specifically, we train one probe for target modality of image and one probe for caption. High probe accuracy reflects that a model is aware of the consistency of the inputs.

Figure \ref{fig:unimodal_probe_accuracy}(b) shows the consistency probe accuracy on Pascal VOC dataset. In all cases, the probe accuracy rises monotonically through a model and approaches around 1.0 in the final layers. Interestingly, the accuracy peak always occurs later than unimodal information has peaked in Figure~\ref{fig:unimodal_probe_accuracy}(a), suggesting a sequential processing of multimodal signals. Overall, these results suggest that VLMs at least in principle have access to information about the (in)consistency between modalities. Admittedly, this does not imply that the VLM uses information about the consistency appropriately, warranting further investigation into how exactly the model integrates, or fails to integrate, these input streams.

\begin{figure}[t]
    \centering
    \includegraphics[width=1\linewidth]{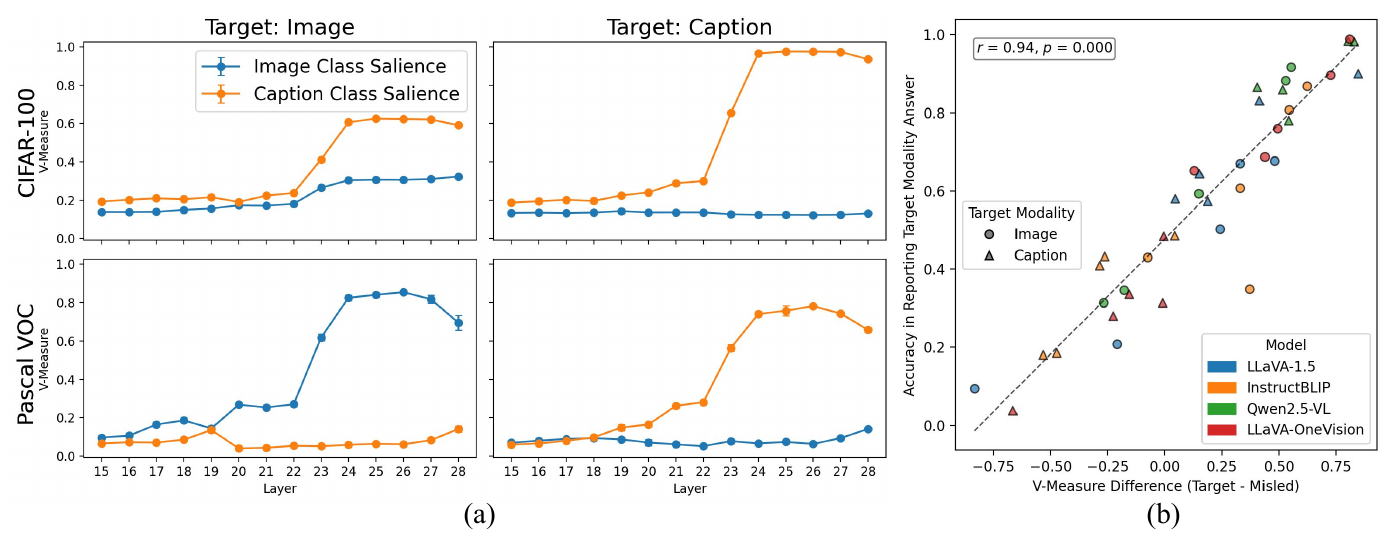}
    \caption{Representational salience and behavioral accuracy. (a) Layer-wise V-Measure of Qwen2.5-VL representations with regard to the image and caption labels of the inconsistent samples, on CIFAR-100 (top) and Pascal VOC (bottom); in CIFAR-100, captions are always encoded more saliently regardless of the target modality, which leads to a degraded performance for Qwen2.5-VL to report image informtion. In contrast, Qwen2.5-VL (re)organizes its representations in Pascal VOC to always align the target modality more saliently.
    (b) Across all model–dataset–task triples, the V-Measure gap (target $-$ misleading) strongly predicts accuracy on the target modality ($r = 0.94$ , $p < 10^{-3}$).}
    \label{fig:clustering_results}
\end{figure}

\subsection{Do VLMs Inherently Favor One Modality over Another in Representational Space?} \label{sec:failure_on_target_modality}

If models always encode information of some modality more saliently than others, it might lead to their degradation when prompted to report information from another less-saliently represented modality. 
To assess which modality is encoded more saliently, we apply K-Means clustering to the hidden representations.

Given a dataset with $|C|$ classes, there are up to $|C|^2 - |C|$ possible inconsistent image-caption pairs. We fit a K-Means model with $|C|$ clusters over all inconsistent samples.  
When the clusters better align with the class labels of one modality, then the information of this modality is more salient in the representational space.
For example, if samples sharing the same image label tend to cluster together, then image information dominates in the representation space.

To quantify this, we use V-Measure \citep{rosenberg-hirschberg-2007-v} to compare the predicted cluster assignments against the true class labels from each modality. A higher V-Measure for one modality indicates that its class information is more saliently represented in the hidden space. 

With the V-Measure for both modalities, we can further compute the V-Measure difference between the target modality and the non-target modality.
If the V-Measure difference is positive, it implies that the model encodes target modality more saliently, and vice versa.  We use the V-Measure difference as a proxy of the relative salience of the target modality, and see if it correlates with the behavioral preference towards the target modality.

Figure \ref{fig:clustering_results}(a) shows V-Measure for Qwen2.5-VL on CIFAR-100 (top) and Pascal VOC (bottom), averaged across three random seeds for intializing the K-Means clustering. We focus on this setting since we can understand why Qwen2.5-VL can perform well on Pascal VOC but not CIFAR datasets.
We observe that when the model performs poorly on CIFAR-100, the model consistently clusters inputs according to the caption, even when the instruction requests information about the image. In contrast, when the model performs well on Pascal VOC, the representational space is dynamically restructured to be consistent with the target modality. This pattern suggests that models which perform poorly fail to (re)-organize their internal representations of the inputs in a way that is responsive to the instruction. For visualizations of V-Measure plots for other models and datasets, see Appendix~\ref{app:representational_salience_plots}.

This ability to (re)cluster the inputs appears to be correlated with performance in general, across models and datasets. Figure~\ref{fig:clustering_results}(b) plots the difference in V-Measure of the target and non-target modality against behavioral accuracy on the target modality for every \textit{model–dataset–modality} triplet, with those hidden activations from the last layer of the models. The strong positive correlation ($r = 0.94$ , $p < 10^{-3}$) confirms that the structure of the representations predicts behavior: models tend to answer from whichever modality their top-layer representations encode most distinctly. 
\section{Role of Attention Head in Conflicting Information Processing} \label{sec:head_attribution}

Inspired by recent work on mediating between sources of factual knowledge \citep{yu-etal-2023-characterizing}, we here ask whether we can find specific components in the model weights that govern the model's choice to report information from the image versus the caption. 

\subsection{Attention Head Intervention}

We look at the impact of each individual attention head on the VLMs' internal processing of conflicting multimodal inputs.
We hypothesize that there exists a competition process when model needs to selectively recall information from one modality over another. We adjust the outputs of each attention head on the last token position of the prompt while keeping all other components the same and observe the performance of the models \footnote{Performance is determined by how well a model can predict the next token, where the next token can just be a subtoken of a class (e.g., ``\_F'' for ``Frog''). During random sampling, we filter out the examples where the answers for image information and caption information share the same first token.}. To adjust the outputs of an attention head, we multiply the output entries by a constant $\alpha$, ranging from -10 to 10, where $\alpha=1$ means the outputs remain the same. We randomly sample 100 conflicting image-caption pairs from Pascal VOC dataset, and generate prompts asking for either image or caption information, resulting in 200 data samples in total. Given that Qwen2.5-VL performs well on most datasets, we focus our analysis on its attention heads and study whether intervening them could make it perform better on the CIFAR datasets.

\begin{figure}[t]
    \centering
    \includegraphics[width=\linewidth]{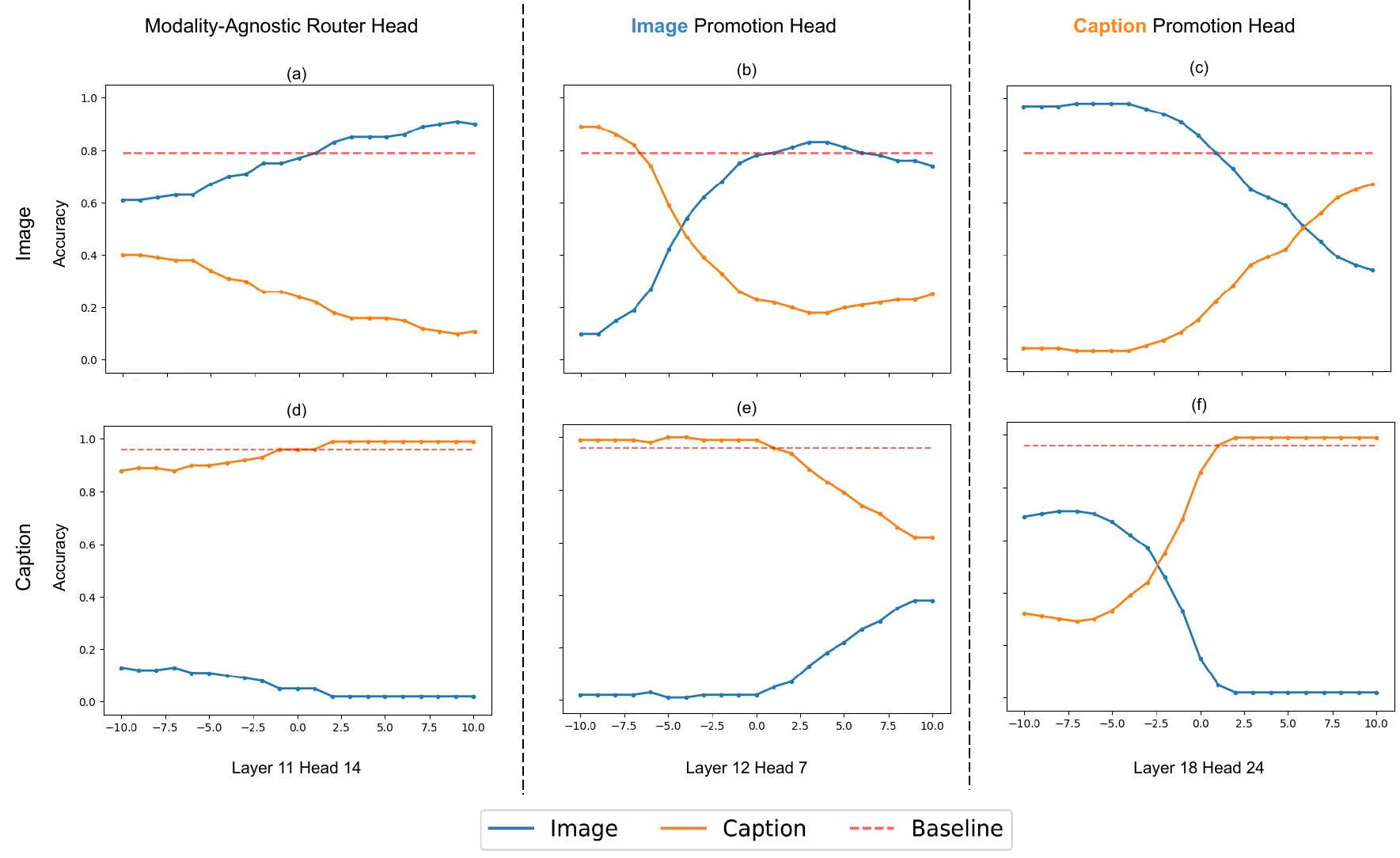}
    \caption{Examples of different types of attention heads in Qwen2.5-VL found on Pascal VOC. First row shows the results on 100 data examples requesting for image information, while second row shows those requesting for caption information. \texttt{Baseline} represents original model's performance without intervention (i.e., $\alpha=1$). After scaling up the attention head outputs by a factor of $\alpha$ (x-axis): (1) \textbf{Modality-agnostic router head} promotes the answers corresponding to the target modality; (2) \textbf{Image promotion head} always promotes the answers about image information; (3) \textbf{Caption promotion head} always promotes the answers about caption information.
    }
    \label{fig:head_types}
\end{figure}

\subsection{Modality-Agnostic Router Head \& Modality-Specific Promotion Head}
\label{sec:head_types}

Based on the trend of model behavior after multiplying an attention head's output by $\alpha$, we discover two interesting types of attention heads in the (Figure \ref{fig:head_types}). See Appendix~\ref{app:classification_of_attention_heads} for a detailed description of how we classify attention heads by their intervention patterns.

\textbf{Modality-agnostic router heads} dynamically promote the answer of a specific modality based on the target modality. When $\alpha$ increases, a router head better promotes the answers of the target modality. When $\alpha$ decreases, model behavior tends to randomly report the answers from either modality. These router heads are interesting since they inherently have a tendency towards target modality (at $\alpha=1$) and so are able to adjust model's behavior dynamically based on the given prompts.

\textbf{Modality-specific promotion heads} promote the answer of a specific modality regardless of the requested modality. When $\alpha$ increases, for example, a caption promotion head will always promote the answers of caption information more frequently. 

In the following sections, we examine the effectiveness of intervening these attention heads on conflicting information processing and whether these heads serve similar functions across datasets. Note that we adopt a subtle difference between how we classify the attention heads and how we determine which attention heads to intervene on for better conflicting information processing capability of VLMs. During classification, we mainly focus on the increasing/decreasing trend of behavioral accuracy when $\alpha$ increases. However, such criterion does not prioritize the \textit{intervenability} of an attention head, which is the \textit{best} performance the model can reach after the intervention on this head. For the following sections, we pick the attention heads which brings significant performance boosts after the intervention.

\subsection{Cross-Dataset Generalization of Router Heads \& Promotion Heads}
\label{sec:head_intervention_generalizability}

We aim to understand the role of attention heads in conflicting information processing in general, and study whether intervening them can alleviate models' performance drop in such context. To this end, we investigate whether the important attention heads found using Pascal VOC data exhibit similar functions upon intervention across other datasets.

Table~\ref{tab:head_generalization_relax} shows the cross-dataset intervention results on the router head at layer 11 head 14 (henceforth L11H14), the image promotion head L19H26, and the caption promotion head L13H26 of Qwen2.5-VL found on Pascal VOC. We fix the multiplicative constant $\alpha$ to 10 here, amplifying the outputs of these attention heads along their original directions. For the modality-agnostic router head, we observe performance improvements regardless of the target modality. This suggests that L11H14 serves a similar function -- i.e., promoting the correct answer across modalities -- across nearly all of the datasets we check. The exception is CUB-Color, which may be caused by Qwen2.5-VL's original suboptimal image unimodal performance CUB-Color \footnote{Though the performance drops on CIFAR-10 and CIFAR-100 when target modality is caption, we attribute this to the saturated performance on these two datasets (98.3\% and 98.4\% respectively).}. For the image promotion head L19H26, we observe performance improvements when the target modality is image while a performance drop when the target modality is caption -- this is consistent with our expectation as the role of image promotion heads is to \textit{invariantly} promote information from the image modality. Conversely, for the caption promotion head L13H26, we observe performance improvements when the target modality is caption while a performance drop when the target modality is image, consistent with our expectation of caption promotion heads.

\begin{table}[]
\centering
\caption{Cross-dataset generalization of modality-agnostic router head (L11H14), image promotion head (L19H26), and caption promotion head (L13H26) of Qwen2.5-VL. $\Delta$ computes the difference between the original model performance and the model performance with an attention head intervened. The router head can generalize across majority of the tasks, except CUB-Color. The image/caption promotion heads invariantly improve model's capability on reporting the corresponding modality's information.
}
\adjustbox{max width=\textwidth}{
    \begin{tabular}{r|rrrrr|rrrrr}
    \toprule
    & \multicolumn{5}{c|}{Target Modality = Image}                                & \multicolumn{5}{c}{Target Modality = Caption}                              \\
    \cmidrule{2-11}
    & Pascal VOC & CIFAR-10 & CIFAR-100 & ImageNet100 & CUB-Color & Pascal & CIFAR-10 & CIFAR-100 & ImageNet100 & CUB-Color \\ 
    \midrule
    Original    & 91.7       & 34.5        & 31.4         & 88.3            & 59.4           & 86.0       & 98.3        & 98.4         & 86.6            & 78.1           \\
    \midrule
    Router Head & 95.2       & 44.5        & 40.6         & 91.3            & 60.8           & 92.3       & 98.1        & 97.5         & 88.3            & 73.2           \\
    $\Delta$       & \cellcolor{comfortgreen} +3.5        & \cellcolor{comfortgreen} +9.9         & \cellcolor{comfortgreen}+9.1          & \cellcolor{comfortgreen}+3.0             & \cellcolor{comfortgreen}+1.4            & \cellcolor{comfortgreen}+6.3        & \cellcolor{mistyrose} -0.2        & \cellcolor{mistyrose}-0.8         & \cellcolor{comfortgreen}+1.6             & \cellcolor{mistyrose}-4.9           \\
    \midrule
Image Head & 96.3                                          & 47.5                                           & 41.5                                           & 92.7                                          & 63.1                                          & 55.0                                           & 90.5                                       & 89.0                                          & 68.0                                          & 43.5                                          \\
$\Delta$            & \cellcolor{comfortgreen}+4.6 & \cellcolor{comfortgreen}+12.9 & \cellcolor{comfortgreen}+10.1 & \cellcolor{comfortgreen}+4.5 & \cellcolor{comfortgreen}+3.7 & \cellcolor{mistyrose}-31.0    & \cellcolor{mistyrose}-7.7 & \cellcolor{mistyrose}-9.4    & \cellcolor{mistyrose}-18.6   & \cellcolor{mistyrose}-34.6   \\
    \midrule
Caption Head & 73.5       & 24.8        & 24.5         & 70.2            & 40.3           & 96.3       & 99.3        & 98.8         & 95.2            & 95.6           \\
    $\Delta$       & \cellcolor{mistyrose}-18.3      & \cellcolor{mistyrose}-9.7        & \cellcolor{mistyrose}-7.0         & \cellcolor{mistyrose}-18.0           & \cellcolor{mistyrose}-19.1          & \cellcolor{comfortgreen}+10.3       & \cellcolor{comfortgreen}+1.0         & \cellcolor{comfortgreen}+0.4          & \cellcolor{comfortgreen}+8.6             & \cellcolor{comfortgreen}+17.5           \\ 
    \bottomrule
    \end{tabular}
}
\label{tab:head_generalization_relax}
\end{table}

\begin{table}[t]
\centering
\caption{Effect of intervening router head (L11H14), image promotion head (L19H26), or caption promotion head (L13H26) on V-Measure difference between the target and non-target modality. $\Delta$ is the difference between the original model and the model with different intervened heads.}

\adjustbox{max width=\textwidth}{
    \begin{tabular}{r|rrrrr|rrrrr}
    \toprule
    & \multicolumn{5}{c|}{Target Modality = Image}                                & \multicolumn{5}{c}{Target Modality = Caption}                              \\
    \cmidrule{2-11}
    & Pascal VOC & CIFAR-10 & CIFAR-100 & ImageNet100 & CUB-Color & Pascal & CIFAR-10 & CIFAR-100 & ImageNet100 & CUB-Color \\ 
    \midrule
    Original & 0.567 & -0.185 & -0.277 & 0.515 & 0.101 & 0.502 & 0.829 & 0.808 & 0.411 & 0.472\\
\midrule
Router Head & 0.771 & -0.033 & -0.123 & 0.608 & 0.170 & 0.624 & 0.835 & 0.771 & 0.434 & 0.518\\
$\Delta$ & \cellcolor{comfortgreen}+0.204 & \cellcolor{comfortgreen}+0.153 & \cellcolor{comfortgreen}+0.154 & \cellcolor{comfortgreen}+0.093 & \cellcolor{comfortgreen}+0.069 & \cellcolor{comfortgreen}+0.122 & \cellcolor{comfortgreen}+0.006 & \cellcolor{mistyrose}-0.037 & \cellcolor{comfortgreen}+0.023 & \cellcolor{comfortgreen}+0.046\\
\midrule
Image Head & 0.684 & -0.016 & -0.083 & 0.612 & 0.181 & 0.227 & 0.557 & 0.547 & 0.166 & 0.094\\
$\Delta$ & \cellcolor{comfortgreen}+0.118 & \cellcolor{comfortgreen}+0.170 & \cellcolor{comfortgreen}+0.194 & \cellcolor{comfortgreen}+0.097 & \cellcolor{comfortgreen}+0.080 & \cellcolor{mistyrose}-0.275 & \cellcolor{mistyrose}-0.272 & \cellcolor{mistyrose}-0.261 & \cellcolor{mistyrose}-0.244 & \cellcolor{mistyrose}-0.378\\
\midrule
Caption Head & 0.174 & -0.348 & -0.366 & 0.256 & 0.003 & 0.747 & 0.937 & 0.841 & 0.614 & 0.922\\
$\Delta$ & \cellcolor{mistyrose}-0.393 & \cellcolor{mistyrose}-0.163 & \cellcolor{mistyrose}-0.089 & \cellcolor{mistyrose}-0.259 & \cellcolor{mistyrose}-0.098 & \cellcolor{comfortgreen}+0.245 & \cellcolor{comfortgreen}+0.108 & \cellcolor{comfortgreen}+0.034 & \cellcolor{comfortgreen}+0.203 & \cellcolor{comfortgreen}+0.450\\

    \bottomrule
    \end{tabular}
}
\label{tab:head_generalization_clustering}
\end{table}

\subsection{Effects of Attention Head Intervention on Representational Salience}
Given that observed effect of the router heads and image/caption promotion heads on model behavior, we ask whether this is due to a direct effect of the heads on the representational structure discussed in Section \ref{sec:failure_on_target_modality} (Figure \ref{fig:clustering_results}).
To test this hypothesis, we run the same clustering analysis as in Section~\ref{sec:failure_on_target_modality} after intervening Qwen2.5-VL on the task-relevant attention heads, and study how these interventions change Qwen2.5-VL's representations of image and caption class information. 

Table~\ref{tab:head_generalization_clustering} shows the V-Measure computed on representations of inconsistent inputs from Qwen2.5-VL before and after intervening the router head L11H14, image head L19H26, and caption head L13H26. We fix $\alpha$ to 10 as in Section~\ref{sec:head_intervention_generalizability}. For router head L11H14, we observe that the differences ($\Delta$) in V-measure of target modality and the non-target modality increase after the intervention across different datasets and modalities, except CIFAR-100, indicating it further helps to make the target modality more saliently represented. We attribute the negative $\Delta$ on CIFAR-100 to the already large V-Measure difference before the intervention. We also notice that though there is a positive $\Delta$ of 0.046 on CUB-Color, the model's performance drops for 4.9\% after the intervention (Table \ref{tab:head_generalization_relax}). This implies that the change in V-Measure is not always perfectly correlated with the change in performance. For image promotion head L19H26, $\Delta$ values are positive when target modality is image, and negative when target modality is caption, indicating that image promotion head always make image information more saliently represented in the hidden space. Conversely, for caption promotion head L13H26, we observe same trend towards the caption modality: caption promotion always promotes the caption information to be more saliently represented in model's hidden space, thus showing a positive change when the target modality is caption and negative when the target modality is image. Comparing across Table \ref{tab:head_generalization_relax} and \ref{tab:head_generalization_clustering}, we observe a positive correlation between the change in V-Measure difference and the change in model's performance. 

\section{Related Work}
\label{sec:related_work}

\paragraph{Conflicting Information Processing in VLMs}
A key challenge in multimodal AI is how models handle conflicting information across modalities, a problem studied by new benchmarks and analyses. Complementary work studies when inconsistencies arise in the knowledge represented in VLMs and investigates the decision-making mechanisms that VLMs use to resolve these inconsistencies \citep{zhu2024unraveling, golovanevsky2025pixels}. Benchmarks, like MMIR (Multimodal Inconsistency Reasoning) \citep{yan2025multimodalinconsistencyreasoningmmir}, further assess how well models can detect and reason about such conflicting multimodal inputs, in more granular setups. In other highly relevant concurrent work, \cite{deng2025wordsvisionvisionlanguagemodels} examines how models respond to queries about image content when the image and its textual description disagree. Their empirical findings indicate that VLMs exhibit a ``blind faith'' in text across various settings. In our work, we not only investigate scenarios where models are prompted to report image content given misleading textual descriptions but also examine the reciprocal situation: how models report textual information when presented inconsistent inputs. Our representational and interventional analyses further complement this existing research.

\paragraph{Interpretability of (Vision-)Language Models}
Interpretability research seeks to understand the inner workings of deep neural networks, primarily focusing on two main approaches: 1) investigating the rich representations learned by these networks, often through decoding methods such as probing \citep{alain2016understanding,tenney2019bert,tenneyyou,belinkov-2022-probing,yun2021does,yun2023emergence}, and 2) revealing the causal and functional roles of model components (e.g., attention heads) in contributing to final predictions \citep{voita-etal-2019-analyzing, NEURIPS2020_92650b2e, wang2022interpretability, hanna2023how, li2023inferencetime, mueller2024quest}. Directly relevant to our work, \citet{yu-etal-2023-characterizing} investigated how language models handle conflicts between their parametric knowledge (stored in weights) and information provided in-context. Their finding that specific attention heads contribute to the models' preference for one information source over another inspires our analogous analysis of how VLMs decide to report information from one modality over another when facing with similar conflicts. More recently, research has begun to adapt interpretability methods, initially developed for unimodal language models, to the vision the multimodal domain \citep{lepori2024beyond, golovanevsky-etal-2025-vlms, neo2025towards, papadimitriou2025interpreting, nikankin2025same}.

\section{Conclusion}

In this work, we demonstrate that VLMs encounter difficulties with straightforward information-reporting tasks when faced with conflicting inputs across modalities. We observe distinct patterns: models such as InstructBLIP and LLaVA-Onevision consistently prefer information from the image modality, whereas models like Qwen2.5-VL demonstrate the capacity to dynamically select the correct modality across multiple datasets. Probing experiments suggest that VLMs effectively encode unimodal information from both modalities and successfully detect whether multimodal inputs are consistent. Furthermore, through representational salience analyses using clustering techniques, we discover a strong correlation between models' behavioral accuracy and the comparative representational salience between the target and non-target modalities. Specifically, models tend to provide correct predictions when their representations more saliently encode the target modality, whereas they tend to produce misled answers when the non-target modality is encoded more saliently.

We further examine the roles of individual attention heads for VLMs in processing conflicting inputs. Through attention output interventions, we identify \textit{modality-agnostic router heads} and \textit{modality-specific promotion heads}. Leveraging these task-relevant attention heads enables us to manipulate the representational salience of target modality and to improve model performance in resolving multimodal conflicts across various datasets. Together, this work provides first steps towards understanding and controlling VLMs in dealing with conflicting information across modalities.

\section*{Acknowledgments}
We thank Ruochen Zhang, Yong Zheng-Xin, Michael Lepori, Aalok Sathe, Apoorv Khandelwal, Jack Merullo, Jacob Russin, Zhuonan Yang, Reza Esfandiarpoor and other members of the Brown Superlab for their valuable feedback on our paper. We would like to thank Calvin Luo for his technical support. The project depicted is sponsored in part by a Young Faculty Award from the Defense Advanced Research Projects Agency, Grant \#D24AP00261. The content of the information does not necessarily reflect the position, or the policy of the government and no official endorsement of this work should be inferred.

\bibliographystyle{plainnat}
\bibliography{custom}

\begin{thebibliography}{41}
\providecommand{\natexlab}[1]{#1}
\providecommand{\url}[1]{\texttt{#1}}
\expandafter\ifx\csname urlstyle\endcsname\relax
  \providecommand{\doi}[1]{doi: #1}\else
  \providecommand{\doi}{doi: \begingroup \urlstyle{rm}\Url}\fi

\bibitem[Acosta et~al.(2022)Acosta, Falcone, Rajpurkar, and Topol]{acosta2022multimodal}
Juli{\'a}n~N Acosta, Guido~J Falcone, Pranav Rajpurkar, and Eric~J Topol.
\newblock Multimodal biomedical ai.
\newblock \emph{Nature medicine}, 28\penalty0 (9):\penalty0 1773--1784, 2022.

\bibitem[Alain and Bengio(2016)]{alain2016understanding}
Guillaume Alain and Yoshua Bengio.
\newblock Understanding intermediate layers using linear classifier probes.
\newblock \emph{arXiv preprint arXiv:1610.01644}, 2016.

\bibitem[Alayrac et~al.(2022)Alayrac, Donahue, Luc, Miech, Barr, Hasson, Lenc, Mensch, Millicah, Reynolds, Ring, Rutherford, Cabi, Han, Gong, Samangooei, Monteiro, Menick, Borgeaud, Brock, Nematzadeh, Sharifzadeh, Binkowski, Barreira, Vinyals, Zisserman, and Simonyan]{10.5555/3600270.3601993}
Jean-Baptiste Alayrac, Jeff Donahue, Pauline Luc, Antoine Miech, Iain Barr, Yana Hasson, Karel Lenc, Arthur Mensch, Katie Millicah, Malcolm Reynolds, Roman Ring, Eliza Rutherford, Serkan Cabi, Tengda Han, Zhitao Gong, Sina Samangooei, Marianne Monteiro, Jacob Menick, Sebastian Borgeaud, Andrew Brock, Aida Nematzadeh, Sahand Sharifzadeh, Mikolaj Binkowski, Ricardo Barreira, Oriol Vinyals, Andrew Zisserman, and Karen Simonyan.
\newblock Flamingo: a visual language model for few-shot learning.
\newblock In \emph{Proceedings of the 36th International Conference on Neural Information Processing Systems}, NIPS '22, Red Hook, NY, USA, 2022. Curran Associates Inc.
\newblock ISBN 9781713871088.

\bibitem[Bai et~al.(2025)Bai, Chen, Liu, Wang, Ge, Song, Dang, Wang, Wang, Tang, Zhong, Zhu, Yang, Li, Wan, Wang, Ding, Fu, Xu, Ye, Zhang, Xie, Cheng, Zhang, Yang, Xu, and Lin]{Qwen2.5-VL}
Shuai Bai, Keqin Chen, Xuejing Liu, Jialin Wang, Wenbin Ge, Sibo Song, Kai Dang, Peng Wang, Shijie Wang, Jun Tang, Humen Zhong, Yuanzhi Zhu, Mingkun Yang, Zhaohai Li, Jianqiang Wan, Pengfei Wang, Wei Ding, Zheren Fu, Yiheng Xu, Jiabo Ye, Xi~Zhang, Tianbao Xie, Zesen Cheng, Hang Zhang, Zhibo Yang, Haiyang Xu, and Junyang Lin.
\newblock Qwen2.5-vl technical report.
\newblock \emph{arXiv preprint arXiv:2502.13923}, 2025.

\bibitem[Belinkov(2022)]{belinkov-2022-probing}
Yonatan Belinkov.
\newblock Probing classifiers: Promises, shortcomings, and advances.
\newblock \emph{Computational Linguistics}, 48\penalty0 (1):\penalty0 207--219, March 2022.
\newblock \doi{10.1162/coli_a_00422}.
\newblock URL \url{https://aclanthology.org/2022.cl-1.7/}.

\bibitem[Cui et~al.(2024)Cui, Aparcedo, Jang, and Lim]{cui2024robustness}
Xuanming Cui, Alejandro Aparcedo, Young~Kyun Jang, and Ser-Nam Lim.
\newblock On the robustness of large multimodal models against image adversarial attacks.
\newblock In \emph{Proceedings of the IEEE/CVF Conference on Computer Vision and Pattern Recognition}, pages 24625--24634, 2024.

\bibitem[Dai et~al.(2023)Dai, Li, Li, Tiong, Zhao, Wang, Li, Fung, and Hoi]{dai2023instructblip}
Wenliang Dai, Junnan Li, Dongxu Li, Anthony Meng~Huat Tiong, Junqi Zhao, Weisheng Wang, Boyang Li, Pascale Fung, and Steven C.~H. Hoi.
\newblock Instructblip: Towards general-purpose vision-language models with instruction tuning.
\newblock In Alice Oh, Tristan Naumann, Amir Globerson, Kate Saenko, Moritz Hardt, and Sergey Levine, editors, \emph{Advances in Neural Information Processing Systems 36: Annual Conference on Neural Information Processing Systems 2023, NeurIPS 2023, New Orleans, LA, USA, December 10 - 16, 2023}, 2023.
\newblock URL \url{http://papers.nips.cc/paper\_files/paper/2023/hash/9a6a435e75419a836fe47ab6793623e6-Abstract-Conference.html}.

\bibitem[Deng et~al.(2025)Deng, Cao, Chen, and Hooi]{deng2025wordsvisionvisionlanguagemodels}
Ailin Deng, Tri Cao, Zhirui Chen, and Bryan Hooi.
\newblock Words or vision: Do vision-language models have blind faith in text?, 2025.
\newblock URL \url{https://arxiv.org/abs/2503.02199}.

\bibitem[Deng et~al.(2009)Deng, Dong, Socher, Li, Li, and Fei-Fei]{deng2009imagenet}
Jia Deng, Wei Dong, Richard Socher, Li-Jia Li, Kai Li, and Li~Fei-Fei.
\newblock Imagenet: A large-scale hierarchical image database.
\newblock In \emph{2009 IEEE conference on computer vision and pattern recognition}, pages 248--255. Ieee, 2009.

\bibitem[Everingham et~al.(2010)Everingham, Van~Gool, Williams, Winn, and Zisserman]{pascal}
Mark Everingham, Luc Van~Gool, Christopher Williams, John Winn, and Andrew Zisserman.
\newblock The pascal visual object classes (voc) challenge.
\newblock \emph{International Journal of Computer Vision}, 88:\penalty0 303--338, 06 2010.
\newblock \doi{10.1007/s11263-009-0275-4}.

\bibitem[Golovanevsky et~al.(2025{\natexlab{a}})Golovanevsky, Rudman, Lepori, Bar, Singh, and Eickhoff]{golovanevsky2025pixels}
Michal Golovanevsky, William Rudman, Michael Lepori, Amir Bar, Ritambhara Singh, and Carsten Eickhoff.
\newblock Pixels versus priors: Controlling knowledge priors in vision-language models through visual counterfacts.
\newblock \emph{arXiv preprint arXiv:2505.17127}, 2025{\natexlab{a}}.

\bibitem[Golovanevsky et~al.(2025{\natexlab{b}})Golovanevsky, Rudman, Palit, Eickhoff, and Singh]{golovanevsky-etal-2025-vlms}
Michal Golovanevsky, William Rudman, Vedant Palit, Carsten Eickhoff, and Ritambhara Singh.
\newblock What do {VLM}s {NOTICE}? a mechanistic interpretability pipeline for {G}aussian-noise-free text-image corruption and evaluation.
\newblock In Luis Chiruzzo, Alan Ritter, and Lu~Wang, editors, \emph{Proceedings of the 2025 Conference of the Nations of the Americas Chapter of the Association for Computational Linguistics: Human Language Technologies (Volume 1: Long Papers)}, pages 11462--11482, Albuquerque, New Mexico, April 2025{\natexlab{b}}. Association for Computational Linguistics.
\newblock ISBN 979-8-89176-189-6.
\newblock URL \url{https://aclanthology.org/2025.naacl-long.571/}.

\bibitem[Hanna et~al.(2023)Hanna, Liu, and Variengien]{hanna2023how}
Michael Hanna, Ollie Liu, and Alexandre Variengien.
\newblock How does {GPT}-2 compute greater-than?: Interpreting mathematical abilities in a pre-trained language model.
\newblock In \emph{Thirty-seventh Conference on Neural Information Processing Systems}, 2023.
\newblock URL \url{https://openreview.net/forum?id=p4PckNQR8k}.

\bibitem[Heinz et~al.(2024)Heinz, Mackin, Trudeau, Bhattacharya, Wang, Banta, Jewett, Salzhauer, Griffin, and Jacobson]{heinz2024evaluating}
Michael~V Heinz, Daniel~M Mackin, Brianna~M Trudeau, Sukanya Bhattacharya, Yinzhou Wang, Haley~A Banta, Abi~D Jewett, Abigail Salzhauer, Tess Griffin, and Nicholas~C Jacobson.
\newblock Evaluating therabot: A randomized control trial investigating the feasibility and effectiveness of a generative ai therapy chatbot for depression, anxiety, and eating disorder symptom treatment.
\newblock \emph{Anxiety, and Eating Disorder Symptom Treatment}, 2024.

\bibitem[Kingma(2014)]{kingma2014adam}
Diederik~P Kingma.
\newblock Adam: A method for stochastic optimization.
\newblock \emph{arXiv preprint arXiv:1412.6980}, 2014.

\bibitem[Krizhevsky et~al.(2009)Krizhevsky, Hinton, et~al.]{krizhevsky2009learning}
Alex Krizhevsky, Geoffrey Hinton, et~al.
\newblock Learning multiple layers of features from tiny images.(2009), 2009.

\bibitem[Lepori et~al.(2024)Lepori, Tartaglini, Vong, Serre, Lake, and Pavlick]{lepori2024beyond}
Michael Lepori, Alexa Tartaglini, Wai~Keen Vong, Thomas Serre, Brenden~M Lake, and Ellie Pavlick.
\newblock Beyond the doors of perception: Vision transformers represent relations between objects.
\newblock \emph{Advances in Neural Information Processing Systems}, 37:\penalty0 131503--131544, 2024.

\bibitem[Li et~al.(2025)Li, Zhang, Guo, Zhang, Li, Zhang, Zhang, Zhang, Li, Liu, and Li]{li2025llavaonevision}
Bo~Li, Yuanhan Zhang, Dong Guo, Renrui Zhang, Feng Li, Hao Zhang, Kaichen Zhang, Peiyuan Zhang, Yanwei Li, Ziwei Liu, and Chunyuan Li.
\newblock {LL}a{VA}-onevision: Easy visual task transfer.
\newblock \emph{Transactions on Machine Learning Research}, 2025.
\newblock ISSN 2835-8856.
\newblock URL \url{https://openreview.net/forum?id=zKv8qULV6n}.

\bibitem[Li et~al.(2023)Li, Patel, Vi{\'e}gas, Pfister, and Wattenberg]{li2023inferencetime}
Kenneth Li, Oam Patel, Fernanda Vi{\'e}gas, Hanspeter Pfister, and Martin Wattenberg.
\newblock Inference-time intervention: Eliciting truthful answers from a language model.
\newblock In \emph{Thirty-seventh Conference on Neural Information Processing Systems}, 2023.
\newblock URL \url{https://openreview.net/forum?id=aLLuYpn83y}.

\bibitem[Liao et~al.(2025)Liao, Lei, Zheng, Moon, Wang, Wang, Paudel, Gool, and Hu]{liao2025benchmarkingmultimodalsemanticsegmentation}
Chenfei Liao, Kaiyu Lei, Xu~Zheng, Junha Moon, Zhixiong Wang, Yixuan Wang, Danda~Pani Paudel, Luc~Van Gool, and Xuming Hu.
\newblock Benchmarking multi-modal semantic segmentation under sensor failures: Missing and noisy modality robustness, 2025.
\newblock URL \url{https://arxiv.org/abs/2503.18445}.

\bibitem[Liu et~al.(2023)Liu, Li, Li, and Lee]{liu2023improvedllava}
Haotian Liu, Chunyuan Li, Yuheng Li, and Yong~Jae Lee.
\newblock Improved baselines with visual instruction tuning, 2023.

\bibitem[Mueller et~al.(2024)Mueller, Brinkmann, Li, Marks, Pal, Prakash, Rager, Sankaranarayanan, Sharma, Sun, et~al.]{mueller2024quest}
Aaron Mueller, Jannik Brinkmann, Millicent Li, Samuel Marks, Koyena Pal, Nikhil Prakash, Can Rager, Aruna Sankaranarayanan, Arnab~Sen Sharma, Jiuding Sun, et~al.
\newblock The quest for the right mediator: A history, survey, and theoretical grounding of causal interpretability.
\newblock \emph{arXiv preprint arXiv:2408.01416}, 2024.

\bibitem[Neo et~al.(2025)Neo, Ong, Torr, Geva, Krueger, and Barez]{neo2025towards}
Clement Neo, Luke Ong, Philip Torr, Mor Geva, David Krueger, and Fazl Barez.
\newblock Towards interpreting visual information processing in vision-language models.
\newblock In \emph{The Thirteenth International Conference on Learning Representations}, 2025.
\newblock URL \url{https://openreview.net/forum?id=chanJGoa7f}.

\bibitem[Nikankin et~al.(2025)Nikankin, Arad, Gandelsman, and Belinkov]{nikankin2025same}
Yaniv Nikankin, Dana Arad, Yossi Gandelsman, and Yonatan Belinkov.
\newblock Same task, different circuits: Disentangling modality-specific mechanisms in vlms.
\newblock \emph{arXiv preprint arXiv:2506.09047}, 2025.

\bibitem[{OpenAI}(2025)]{openai_operator_2025}
{OpenAI}.
\newblock Introducing operator.
\newblock \url{https://openai.com/index/introducing-operator/}, January 2025.
\newblock Published January 23, 2025; accessed 2025-05-14.

\bibitem[OpenAI et~al.(2024)OpenAI, Achiam, Adler, Agarwal, Ahmad, Akkaya, Aleman, Almeida, Altenschmidt, Altman, Anadkat, Avila, Babuschkin, Balaji, Balcom, Baltescu, Bao, Bavarian, Belgum, Bello, Berdine, Bernadett-Shapiro, Berner, Bogdonoff, Boiko, Boyd, Brakman, Brockman, Brooks, Brundage, Button, Cai, Campbell, Cann, Carey, Carlson, Carmichael, Chan, Chang, Chantzis, Chen, Chen, Chen, Chen, Chen, Chess, Cho, Chu, Chung, Cummings, Currier, Dai, Decareaux, Degry, Deutsch, Deville, Dhar, Dohan, Dowling, Dunning, Ecoffet, Eleti, Eloundou, Farhi, Fedus, Felix, Fishman, Forte, Fulford, Gao, Georges, Gibson, Goel, Gogineni, Goh, Gontijo-Lopes, Gordon, Grafstein, Gray, Greene, Gross, Gu, Guo, Hallacy, Han, Harris, He, Heaton, Heidecke, Hesse, Hickey, Hickey, Hoeschele, Houghton, Hsu, Hu, Hu, Huizinga, Jain, Jain, Jang, Jiang, Jiang, Jin, Jin, Jomoto, Jonn, Jun, Kaftan, Łukasz Kaiser, Kamali, Kanitscheider, Keskar, Khan, Kilpatrick, Kim, Kim, Kim, Kirchner, Kiros, Knight, Kokotajlo, Łukasz Kondraciuk, Kondrich,
  Konstantinidis, Kosic, Krueger, Kuo, Lampe, Lan, Lee, Leike, Leung, Levy, Li, Lim, Lin, Lin, Litwin, Lopez, Lowe, Lue, Makanju, Malfacini, Manning, Markov, Markovski, Martin, Mayer, Mayne, McGrew, McKinney, McLeavey, McMillan, McNeil, Medina, Mehta, Menick, Metz, Mishchenko, Mishkin, Monaco, Morikawa, Mossing, Mu, Murati, Murk, Mély, Nair, Nakano, Nayak, Neelakantan, Ngo, Noh, Ouyang, O'Keefe, Pachocki, Paino, Palermo, Pantuliano, Parascandolo, Parish, Parparita, Passos, Pavlov, Peng, Perelman, de~Avila Belbute~Peres, Petrov, de~Oliveira~Pinto, Michael, Pokorny, Pokrass, Pong, Powell, Power, Power, Proehl, Puri, Radford, Rae, Ramesh, Raymond, Real, Rimbach, Ross, Rotsted, Roussez, Ryder, Saltarelli, Sanders, Santurkar, Sastry, Schmidt, Schnurr, Schulman, Selsam, Sheppard, Sherbakov, Shieh, Shoker, Shyam, Sidor, Sigler, Simens, Sitkin, Slama, Sohl, Sokolowsky, Song, Staudacher, Such, Summers, Sutskever, Tang, Tezak, Thompson, Tillet, Tootoonchian, Tseng, Tuggle, Turley, Tworek, Uribe, Vallone, Vijayvergiya,
  Voss, Wainwright, Wang, Wang, Wang, Ward, Wei, Weinmann, Welihinda, Welinder, Weng, Weng, Wiethoff, Willner, Winter, Wolrich, Wong, Workman, Wu, Wu, Wu, Xiao, Xu, Yoo, Yu, Yuan, Zaremba, Zellers, Zhang, Zhang, Zhao, Zheng, Zhuang, Zhuk, and Zoph]{openai2024gpt4technicalreport}
OpenAI, Josh Achiam, Steven Adler, Sandhini Agarwal, Lama Ahmad, Ilge Akkaya, Florencia~Leoni Aleman, Diogo Almeida, Janko Altenschmidt, Sam Altman, Shyamal Anadkat, Red Avila, Igor Babuschkin, Suchir Balaji, Valerie Balcom, Paul Baltescu, Haiming Bao, Mohammad Bavarian, Jeff Belgum, Irwan Bello, Jake Berdine, Gabriel Bernadett-Shapiro, Christopher Berner, Lenny Bogdonoff, Oleg Boiko, Madelaine Boyd, Anna-Luisa Brakman, Greg Brockman, Tim Brooks, Miles Brundage, Kevin Button, Trevor Cai, Rosie Campbell, Andrew Cann, Brittany Carey, Chelsea Carlson, Rory Carmichael, Brooke Chan, Che Chang, Fotis Chantzis, Derek Chen, Sully Chen, Ruby Chen, Jason Chen, Mark Chen, Ben Chess, Chester Cho, Casey Chu, Hyung~Won Chung, Dave Cummings, Jeremiah Currier, Yunxing Dai, Cory Decareaux, Thomas Degry, Noah Deutsch, Damien Deville, Arka Dhar, David Dohan, Steve Dowling, Sheila Dunning, Adrien Ecoffet, Atty Eleti, Tyna Eloundou, David Farhi, Liam Fedus, Niko Felix, Simón~Posada Fishman, Juston Forte, Isabella Fulford, Leo
  Gao, Elie Georges, Christian Gibson, Vik Goel, Tarun Gogineni, Gabriel Goh, Rapha Gontijo-Lopes, Jonathan Gordon, Morgan Grafstein, Scott Gray, Ryan Greene, Joshua Gross, Shixiang~Shane Gu, Yufei Guo, Chris Hallacy, Jesse Han, Jeff Harris, Yuchen He, Mike Heaton, Johannes Heidecke, Chris Hesse, Alan Hickey, Wade Hickey, Peter Hoeschele, Brandon Houghton, Kenny Hsu, Shengli Hu, Xin Hu, Joost Huizinga, Shantanu Jain, Shawn Jain, Joanne Jang, Angela Jiang, Roger Jiang, Haozhun Jin, Denny Jin, Shino Jomoto, Billie Jonn, Heewoo Jun, Tomer Kaftan, Łukasz Kaiser, Ali Kamali, Ingmar Kanitscheider, Nitish~Shirish Keskar, Tabarak Khan, Logan Kilpatrick, Jong~Wook Kim, Christina Kim, Yongjik Kim, Jan~Hendrik Kirchner, Jamie Kiros, Matt Knight, Daniel Kokotajlo, Łukasz Kondraciuk, Andrew Kondrich, Aris Konstantinidis, Kyle Kosic, Gretchen Krueger, Vishal Kuo, Michael Lampe, Ikai Lan, Teddy Lee, Jan Leike, Jade Leung, Daniel Levy, Chak~Ming Li, Rachel Lim, Molly Lin, Stephanie Lin, Mateusz Litwin, Theresa Lopez, Ryan
  Lowe, Patricia Lue, Anna Makanju, Kim Malfacini, Sam Manning, Todor Markov, Yaniv Markovski, Bianca Martin, Katie Mayer, Andrew Mayne, Bob McGrew, Scott~Mayer McKinney, Christine McLeavey, Paul McMillan, Jake McNeil, David Medina, Aalok Mehta, Jacob Menick, Luke Metz, Andrey Mishchenko, Pamela Mishkin, Vinnie Monaco, Evan Morikawa, Daniel Mossing, Tong Mu, Mira Murati, Oleg Murk, David Mély, Ashvin Nair, Reiichiro Nakano, Rajeev Nayak, Arvind Neelakantan, Richard Ngo, Hyeonwoo Noh, Long Ouyang, Cullen O'Keefe, Jakub Pachocki, Alex Paino, Joe Palermo, Ashley Pantuliano, Giambattista Parascandolo, Joel Parish, Emy Parparita, Alex Passos, Mikhail Pavlov, Andrew Peng, Adam Perelman, Filipe de~Avila Belbute~Peres, Michael Petrov, Henrique~Ponde de~Oliveira~Pinto, Michael, Pokorny, Michelle Pokrass, Vitchyr~H. Pong, Tolly Powell, Alethea Power, Boris Power, Elizabeth Proehl, Raul Puri, Alec Radford, Jack Rae, Aditya Ramesh, Cameron Raymond, Francis Real, Kendra Rimbach, Carl Ross, Bob Rotsted, Henri Roussez,
  Nick Ryder, Mario Saltarelli, Ted Sanders, Shibani Santurkar, Girish Sastry, Heather Schmidt, David Schnurr, John Schulman, Daniel Selsam, Kyla Sheppard, Toki Sherbakov, Jessica Shieh, Sarah Shoker, Pranav Shyam, Szymon Sidor, Eric Sigler, Maddie Simens, Jordan Sitkin, Katarina Slama, Ian Sohl, Benjamin Sokolowsky, Yang Song, Natalie Staudacher, Felipe~Petroski Such, Natalie Summers, Ilya Sutskever, Jie Tang, Nikolas Tezak, Madeleine~B. Thompson, Phil Tillet, Amin Tootoonchian, Elizabeth Tseng, Preston Tuggle, Nick Turley, Jerry Tworek, Juan Felipe~Cerón Uribe, Andrea Vallone, Arun Vijayvergiya, Chelsea Voss, Carroll Wainwright, Justin~Jay Wang, Alvin Wang, Ben Wang, Jonathan Ward, Jason Wei, CJ~Weinmann, Akila Welihinda, Peter Welinder, Jiayi Weng, Lilian Weng, Matt Wiethoff, Dave Willner, Clemens Winter, Samuel Wolrich, Hannah Wong, Lauren Workman, Sherwin Wu, Jeff Wu, Michael Wu, Kai Xiao, Tao Xu, Sarah Yoo, Kevin Yu, Qiming Yuan, Wojciech Zaremba, Rowan Zellers, Chong Zhang, Marvin Zhang, Shengjia
  Zhao, Tianhao Zheng, Juntang Zhuang, William Zhuk, and Barret Zoph.
\newblock Gpt-4 technical report, 2024.
\newblock URL \url{https://arxiv.org/abs/2303.08774}.

\bibitem[Papadimitriou et~al.(2025)Papadimitriou, Su, Fel, Saphra, Kakade, and Gil]{papadimitriou2025interpreting}
Isabel Papadimitriou, Huangyuan Su, Thomas Fel, Naomi Saphra, Sham Kakade, and Stephanie Gil.
\newblock Interpreting the linear structure of vision-language model embedding spaces.
\newblock \emph{arXiv preprint arXiv:2504.11695}, 2025.

\bibitem[Rosenberg and Hirschberg(2007)]{rosenberg-hirschberg-2007-v}
Andrew Rosenberg and Julia Hirschberg.
\newblock {V}-measure: A conditional entropy-based external cluster evaluation measure.
\newblock In Jason Eisner, editor, \emph{Proceedings of the 2007 Joint Conference on Empirical Methods in Natural Language Processing and Computational Natural Language Learning ({EMNLP}-{C}o{NLL})}, pages 410--420, Prague, Czech Republic, June 2007. Association for Computational Linguistics.
\newblock URL \url{https://aclanthology.org/D07-1043/}.

\bibitem[Schlarmann and Hein(2023)]{10350690}
Christian Schlarmann and Matthias Hein.
\newblock { On the Adversarial Robustness of Multi-Modal Foundation Models }.
\newblock In \emph{2023 IEEE/CVF International Conference on Computer Vision Workshops (ICCVW)}, pages 3679--3687, Los Alamitos, CA, USA, October 2023. IEEE Computer Society.
\newblock \doi{10.1109/ICCVW60793.2023.00395}.
\newblock URL \url{https://doi.ieeecomputersociety.org/10.1109/ICCVW60793.2023.00395}.

\bibitem[Tenney et~al.()Tenney, Xia, Chen, Wang, Poliak, McCoy, Kim, Van~Durme, Bowman, Das, et~al.]{tenneyyou}
Ian Tenney, Patrick Xia, Berlin Chen, Alex Wang, Adam Poliak, R~Thomas McCoy, Najoung Kim, Benjamin Van~Durme, Samuel~R Bowman, Dipanjan Das, et~al.
\newblock What do you learn from context? probing for sentence structure in contextualized word representations.
\newblock In \emph{International Conference on Learning Representations}.

\bibitem[Tenney et~al.(2019)Tenney, Das, and Pavlick]{tenney2019bert}
Ian Tenney, Dipanjan Das, and Ellie Pavlick.
\newblock Bert rediscovers the classical nlp pipeline.
\newblock In \emph{Proceedings of the 57th Annual Meeting of the Association for Computational Linguistics}, pages 4593--4601, 2019.

\bibitem[Vig et~al.(2020)Vig, Gehrmann, Belinkov, Qian, Nevo, Singer, and Shieber]{NEURIPS2020_92650b2e}
Jesse Vig, Sebastian Gehrmann, Yonatan Belinkov, Sharon Qian, Daniel Nevo, Yaron Singer, and Stuart Shieber.
\newblock Investigating gender bias in language models using causal mediation analysis.
\newblock In H.~Larochelle, M.~Ranzato, R.~Hadsell, M.F. Balcan, and H.~Lin, editors, \emph{Advances in Neural Information Processing Systems}, volume~33, pages 12388--12401. Curran Associates, Inc., 2020.
\newblock URL \url{https://proceedings.neurips.cc/paper_files/paper/2020/file/92650b2e92217715fe312e6fa7b90d82-Paper.pdf}.

\bibitem[Voita et~al.(2019)Voita, Talbot, Moiseev, Sennrich, and Titov]{voita-etal-2019-analyzing}
Elena Voita, David Talbot, Fedor Moiseev, Rico Sennrich, and Ivan Titov.
\newblock Analyzing multi-head self-attention: Specialized heads do the heavy lifting, the rest can be pruned.
\newblock In Anna Korhonen, David Traum, and Llu{\'i}s M{\`a}rquez, editors, \emph{Proceedings of the 57th Annual Meeting of the Association for Computational Linguistics}, pages 5797--5808, Florence, Italy, July 2019. Association for Computational Linguistics.
\newblock \doi{10.18653/v1/P19-1580}.
\newblock URL \url{https://aclanthology.org/P19-1580/}.

\bibitem[Wah et~al.(2011)Wah, Branson, Welinder, Perona, and Belongie]{wah_branson_welinder_perona_belongie_2011}
Catherine Wah, Steve Branson, Peter Welinder, Pietro Perona, and Serge Belongie.
\newblock The caltech-ucsd birds-200-2011 dataset.
\newblock Jul 2011.

\bibitem[Wang et~al.(2022)Wang, Variengien, Conmy, Shlegeris, and Steinhardt]{wang2022interpretability}
Kevin Wang, Alexandre Variengien, Arthur Conmy, Buck Shlegeris, and Jacob Steinhardt.
\newblock Interpretability in the wild: a circuit for indirect object identification in gpt-2 small.
\newblock \emph{arXiv preprint arXiv:2211.00593}, 2022.

\bibitem[Yan et~al.(2025)Yan, Fan, Li, Jiang, Zhao, Guan, Kuo, and Wang]{yan2025multimodalinconsistencyreasoningmmir}
Qianqi Yan, Yue Fan, Hongquan Li, Shan Jiang, Yang Zhao, Xinze Guan, Ching-Chen Kuo, and Xin~Eric Wang.
\newblock Multimodal inconsistency reasoning (mmir): A new benchmark for multimodal reasoning models, 2025.
\newblock URL \url{https://arxiv.org/abs/2502.16033}.

\bibitem[Yu et~al.(2023)Yu, Merullo, and Pavlick]{yu-etal-2023-characterizing}
Qinan Yu, Jack Merullo, and Ellie Pavlick.
\newblock Characterizing mechanisms for factual recall in language models.
\newblock In Houda Bouamor, Juan Pino, and Kalika Bali, editors, \emph{Proceedings of the 2023 Conference on Empirical Methods in Natural Language Processing}, pages 9924--9959, Singapore, December 2023. Association for Computational Linguistics.
\newblock \doi{10.18653/v1/2023.emnlp-main.615}.
\newblock URL \url{https://aclanthology.org/2023.emnlp-main.615/}.

\bibitem[Yun et~al.(2021)Yun, Sun, and Pavlick]{yun2021does}
Tian Yun, Chen Sun, and Ellie Pavlick.
\newblock Does vision-and-language pretraining improve lexical grounding?
\newblock In \emph{Findings of the Association for Computational Linguistics: EMNLP 2021}, pages 4357--4366, 2021.

\bibitem[Yun et~al.(2023)Yun, Zeng, Handa, Thapliyal, Pang, Pavlick, and Sun]{yun2023emergence}
Tian Yun, Zilai Zeng, Kunal Handa, Ashish Thapliyal, Bo~Pang, Ellie Pavlick, and Chen Sun.
\newblock Emergence of abstract state representations in embodied sequence modeling.
\newblock In \emph{Proceedings of the 2023 Conference on Empirical Methods in Natural Language Processing}, pages 12190--12205, 2023.

\bibitem[Zhu et~al.(2025)Zhu, Qi, Gui, Chen, Ngo, and Lim]{zhu2025callingspadeheartgaslighting}
Bin Zhu, Huiyan Qi, Yinxuan Gui, Jingjing Chen, Chong-Wah Ngo, and Ee-Peng Lim.
\newblock Calling a spade a heart: Gaslighting multimodal large language models via negation, 2025.
\newblock URL \url{https://arxiv.org/abs/2501.19017}.

\bibitem[Zhu et~al.(2024)Zhu, Liu, Wang, Tu, and Chen]{zhu2024unraveling}
Tinghui Zhu, Qin Liu, Fei Wang, Zhengzhong Tu, and Muhao Chen.
\newblock Unraveling cross-modality knowledge conflict in large vision-language models.
\newblock \emph{arXiv preprint arXiv:2410.03659}, 2024.

\end{thebibliography}

\tableofcontents
\clearpage

\appendix
\counterwithin{figure}{section}
\counterwithin{table}{section}

\section*{Appendix}

\section{Limitations} 
\label{app:limitations}
Our work aim at studying how vision-language models (VLMs) process conflicting information across modalities. In our experiments, we mainly focus on multimodal large language model (MLLMs), which is the most popular and performant VLMs, but they only represent a subset of the existing VLMs. We also mainly focus on the 7b-parameter models, but there are MLLMs with varying sizes. Thus, it is possible that models with different sizes may behave in different ways or vary in the inherent preferred modalities. For datasets, we consider image classification and object/attribute recognition datasets. Though we show that the attention heads found on these datasets are generalizable to others, it is not guaranteed that those heads can further generalize to other datasets. On that note, we did not run experiments when multimodal data is collected by real-world sensors from different sources, and we hope future works could extend our findings to inputs from modalities beyond text and vision (i.e. audio, spatial, and etc.) Lastly, for our attention head intervention studies, we focus on analyzing the role of individual attention heads in promoting predictions about a specific modality. An interesting future work could explore the interactions among attention heads (i.e. circuit analysis) to get a better understanding of how other components (e.g., multiple heads, MLP layers) together process conflicting information from the inputs.

\section{Experiments Compute Resources}
\label{app:compute_resources}
In all our experiments, we use 60 Nvidia GeForce RTX 3090 (24GB), 8 A100 (80GB), and 4 H100 (80GB). We run majority of the experiments on 2 GPUs at a time. For the behavioral evaluations (Section~\ref{sec:behavioral_results}), it takes about 30 minutes to run through the test set for each model and dataset and experiment setup. For probe training and clustering analysis (Section~\ref{sec:representational_analysis}), it takes up to 10 hours to compute the hidden activations and up to 1 hours for running linear probe training and K-Means clustering, for each dataset-model-target-modality triple. The process of identifying important attention heads of each model can take 0.5 day (for InstructBLIP) to 1.5 days (for LLaVA-OneVision).

\section{Implementation Details}
\label{app:implementation_details}

\subsection{Dataset Preprocessing}
\label{app:data_preprocessing}
For Pascal VOC, we use the 2012 version and filter out all images with more than one object detection annotated with it, for both the train and validation splits. The resulting images all have a single object detection annotation with them and we use the class of that single object as the corresponding class label for each image. For the CUB dataset, we use the \texttt{CUB\_200\_2011} version and filter out all images with more than one label associated with the ``primary color'' attribute. The resulting images all have a single primary-color annotation and we use that as the corresponding class label for each image. Table~\ref{tab:datasets} shows the details of each dataset after preprocessing.

\begin{table}[h]
\centering
\caption{Summary of datasets used in our experiments.}
\begin{tabular}{lccc}
\toprule
\textbf{Dataset} & \textbf{\#Train} & \textbf{\#Test} & \textbf{\#Classes} \\
\midrule
CIFAR-10 & 50K  & 10K & 10 \\
Pascal VOC & 2,470 & 2,486 & 20 \\
CIFAR-100 & 50K & 10K & 100 \\
ImageNet100  & 130K  & 5K  & 100 \\
CUB (Color Attr.) & 2,719 & 2,582 & 15 \\
\bottomrule
\end{tabular}
\label{tab:datasets}
\end{table}

\subsection{Prompt Construction}
\label{app:prompt_templates}

To construct the prompt for the models, we use the following prompt templates for each part of the prompt. The overall structure of the final prompt is: first, we have the image part, where for each model we use their corresponding image token as the placeholder for where the image embeddings will be inserted in; next, we append the caption template with a class label filled in; after that, we append the question template, depending on the target modality; then, we provide the options with the option template and five options that always include the image and class labels (if available); finally, we add on the instruction template. For all the models, we use their respective instruction-template and add the answer template right after the assistant's turn of conversation. Table ~\ref{tab:ic-templates} and ~\ref{tab:cub-templates} shows the prompt templates we used for the image classification datasets and for CUB-Color respectively. Table ~\ref{tab:sample-prompts} includes some generated sample prompts for each of the four models on CIFAR-10.

\begin{table}[h]
  \caption{Templates for Image Classification Datasets}
  \label{tab:ic-templates}
  \centering
  \begin{tabular}{ll}
    \toprule
    \multicolumn{2}{l}{\bfseries Caption templates} \\
    \midrule
    & This is an image of a \texttt{\{CLASS\_LABEL\}}. \\
    & This is a photo of a \texttt{\{CLASS\_LABEL\}}. \\
    & An image of a \texttt{\{CLASS\_LABEL\}}. \\
    & A photo of a \texttt{\{CLASS\_LABEL\}}. \\
    & This is a \texttt{\{CLASS\_LABEL\}}. \\
    & A \texttt{\{CLASS\_LABEL\}}. \\
    \midrule
    \multicolumn{2}{l}{\bfseries Question templates} \\
    \midrule
    & What is mentioned by the caption? \\
    & What does the caption say? \\
    & What is the class indicated by the caption? \\
    & What is the class of the input image? \\
    & What is in the image? \\
    \midrule
    \multicolumn{2}{l}{\bfseries Option template} \\
    \midrule
    & Select from the following classes: \\
    \midrule
    \multicolumn{2}{l}{\bfseries Instruction template} \\
    \midrule
    & Answer the question using a single word or phrase. \\
    \midrule
    \multicolumn{2}{l}{\bfseries Answer template} \\
    \midrule
    & Answer: \\
    \bottomrule
  \end{tabular}
\end{table}

\begin{table}[h]
  \caption{Templates for CUB‑Color Dataset}
  \label{tab:cub-templates}
  \centering
  \begin{tabular}{ll}
    \toprule
    \multicolumn{2}{l}{\bfseries Caption templates} \\
    \midrule
    & This is a \texttt{\{COLOR\_LABEL\}} bird. \\
    & This bird has a primary color of \texttt{\{COLOR\_LABEL\}}. \\
    \midrule
    \multicolumn{2}{l}{\bfseries Question templates} \\
    \midrule
    & What is the primary color of the bird described in the caption? \\
    & What color is the bird described in the caption? \\
    & What is the class indicated by the caption? \\
    & What is the primary color of the bird shown in the image? \\
    & What color is the bird in the image? \\
    \midrule
    \multicolumn{2}{l}{\bfseries Option template} \\
    \midrule
    & Select from the following colors: \\
    \midrule
    \multicolumn{2}{l}{\bfseries Instruction template} \\
    \midrule
    & Answer the question using a single word or phrase. \\
    \midrule
    \multicolumn{2}{l}{\bfseries Answer template} \\
    \midrule
    & Answer: \\
    \bottomrule
  \end{tabular}
\end{table}

\begin{table}[h]
  \caption{Sample prompts for different models on CIFAR-10}
  \label{tab:sample-prompts}
  \centering
  \adjustbox{max width=\linewidth}{
  \begin{tabular}{@{}l p{12cm}@{}}
    \toprule
    \textbf{Model} & \textbf{Prompt} \\
    \midrule
    LLaVA-1.5 &
      \ttfamily
      USER: Image: <image> \newline
      Caption: This is an image of a deer. \newline
      Question: What is in the image? Select from the following classes: deer, horse, bird, automobile, cat. Answer the question using a single word or phrase. \newline
      ASSISTANT: Answer:
      \normalfont
      \\
    \midrule
    InstructBLIP &
      \ttfamily
      USER: Caption: This is an image of a truck. \newline
      Question: What is the class of the input image? Select from the following classes: cat, deer, truck, frog, airplane. Answer the question using a single word or phrase. \newline
      ASSISTANT: Answer:
      \normalfont
      \\
    \midrule
    Qwen2.5-VL &
      \ttfamily
      <|im\_start|>system You are a helpful assistant.<|im\_end|> \newline
      <|im\_start|>user Image: <|vision\_start|><|image\_pad|><|vision\_end|> \newline
      Caption: An image of a truck. Question: What is the class of the input image? Select from the following classes: truck, cat, ship, frog, deer. Answer the question using a single word or phrase. \newline
      <|im\_end|> <|im\_start|>assistant Answer:
      \normalfont
      \\
    \midrule
    LLaVA-OneVision &
      \ttfamily
      <|im\_start|>user Image: <image> \newline
      Caption: This is an image of a dog. Question: What is the class of the input image? Select from the following classes: bird, deer, cat, ship, dog. Answer the question using a single word or phrase. \newline
      <|im\_end|> <|im\_start|>assistant Answer:
      \normalfont
      \\
    \bottomrule
  \end{tabular}
  }
\end{table}

\subsection{Probe Training Configurations} 
\label{app:probe_configuration}

For probe training, we train them for 1,000 epochs and keep track of the model weights that lead to the lowest validation loss. After that, we retrieve the best model weight and run test evaluation on it.

\begin{table}[h]
  \caption{Probe Training Hyperparameters}
  \label{sample-table}
  \centering
  \begin{tabular}{ll}
    \toprule
    ~     & Description     \\
    \midrule
    \#Epochs & 1,000  \\
    Batch Size  & 256  \\
    Learning Rate  & 0.001  \\
    Train-Val Split  & 80\% (Train); 20\% (Validation)  \\
    Optimizer   & Adam  \citep{kingma2014adam}     \\
    \bottomrule
  \end{tabular}
\end{table}

\clearpage

\section{Additional Behavioral Results}
\label{app:behavioral_results}

\subsection{Experiments with Consistent Inputs Across Modalities}
\label{app:consistent_baseline}
\begin{figure}[h]
    \centering
    \includegraphics[width=1\linewidth]{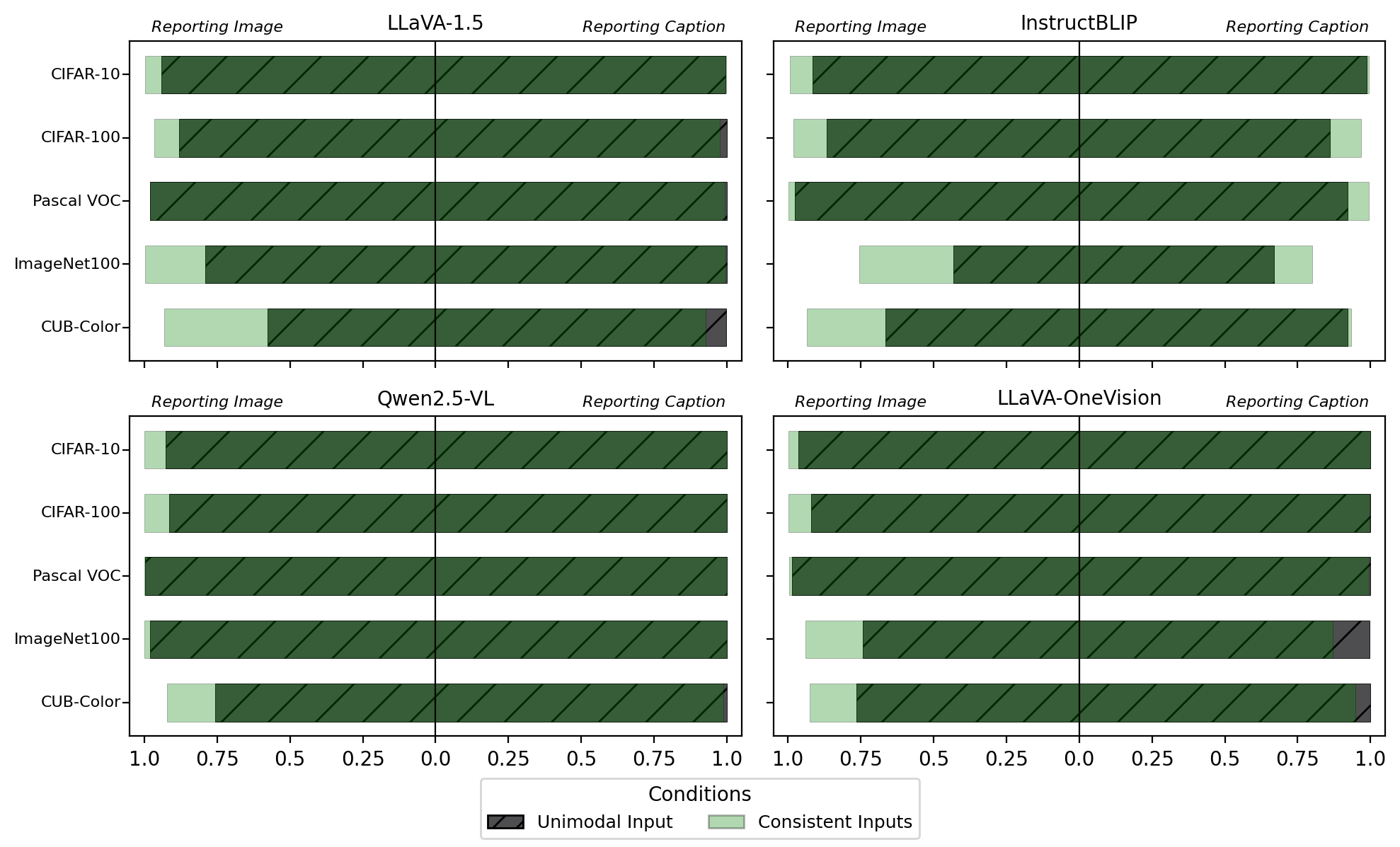}
    \caption{Model performance on reporting target modality under single-modality inputs and consistent inputs. All models acorss nearly all datasets appear to benefit from the complementary information provided by consistent information from the non-target modality.}
    \label{fig:behavior_consistent}
\end{figure}

In addition to unimodal information baselines, we also consider consistent-input experiment, where models are presented with consistent pairs of image and caption. This setting serves as an upper bound, allowing us to estimate the model’s optimal multimodal performance when the multimodal inputs agree with one another. Results are shown in Figure~\ref{fig:behavior_consistent}.

\subsection{Classification of Behavioral Responses}
\label{app:classification_of_behavioral_responses}
\begin{figure}[p]
    \centering
    \includegraphics[width=1.0\linewidth]{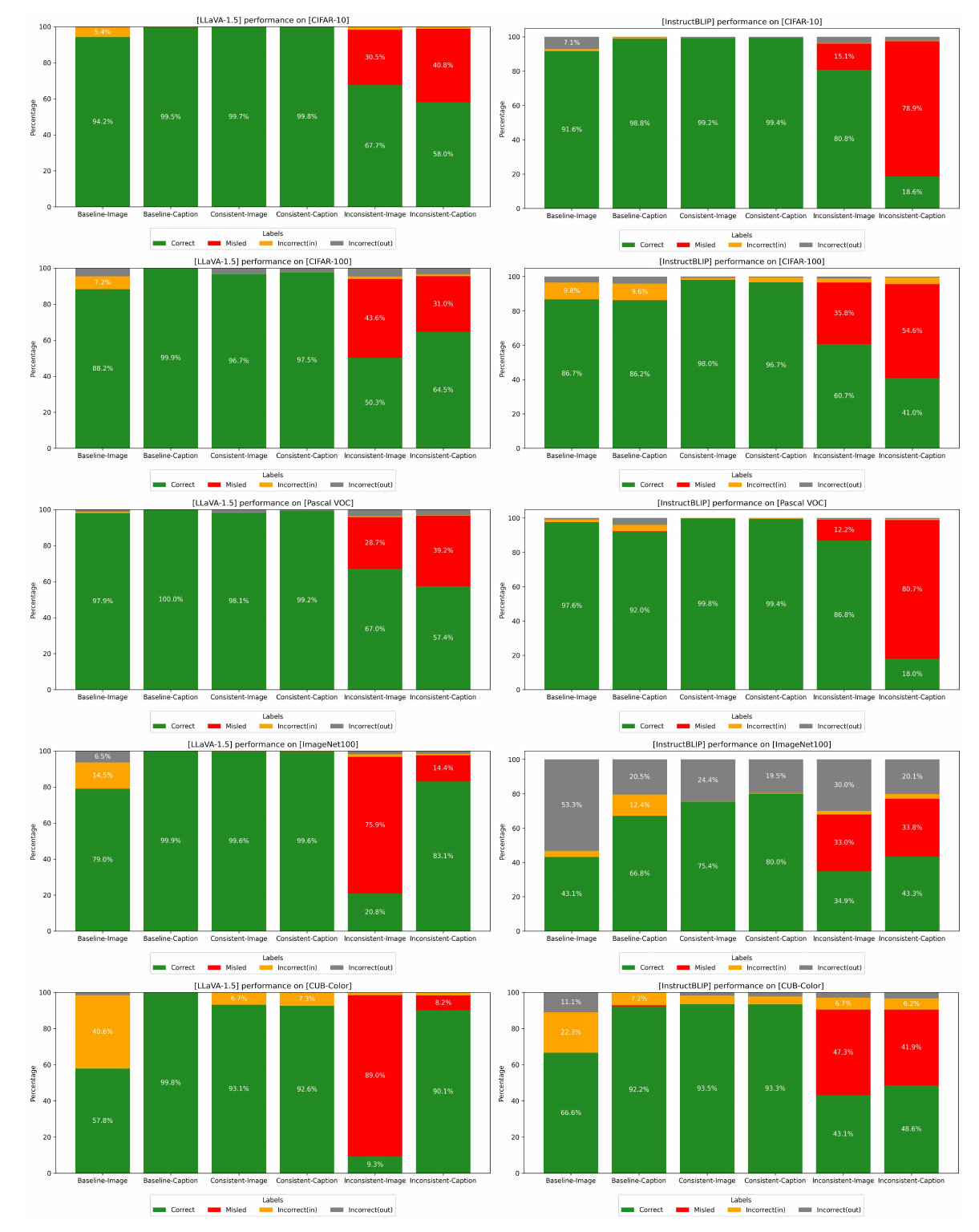}
    \caption{Breakdown of model predictions under unimodal baselines, consistent input pairs and inconsistent input pairs. Left: LLaVA-1.5; right: InstructBLIP.}
    \label{fig:breakdown_llava_blip}
\end{figure}

\begin{figure}[p]
    \centering
    \includegraphics[width=1.0\linewidth]{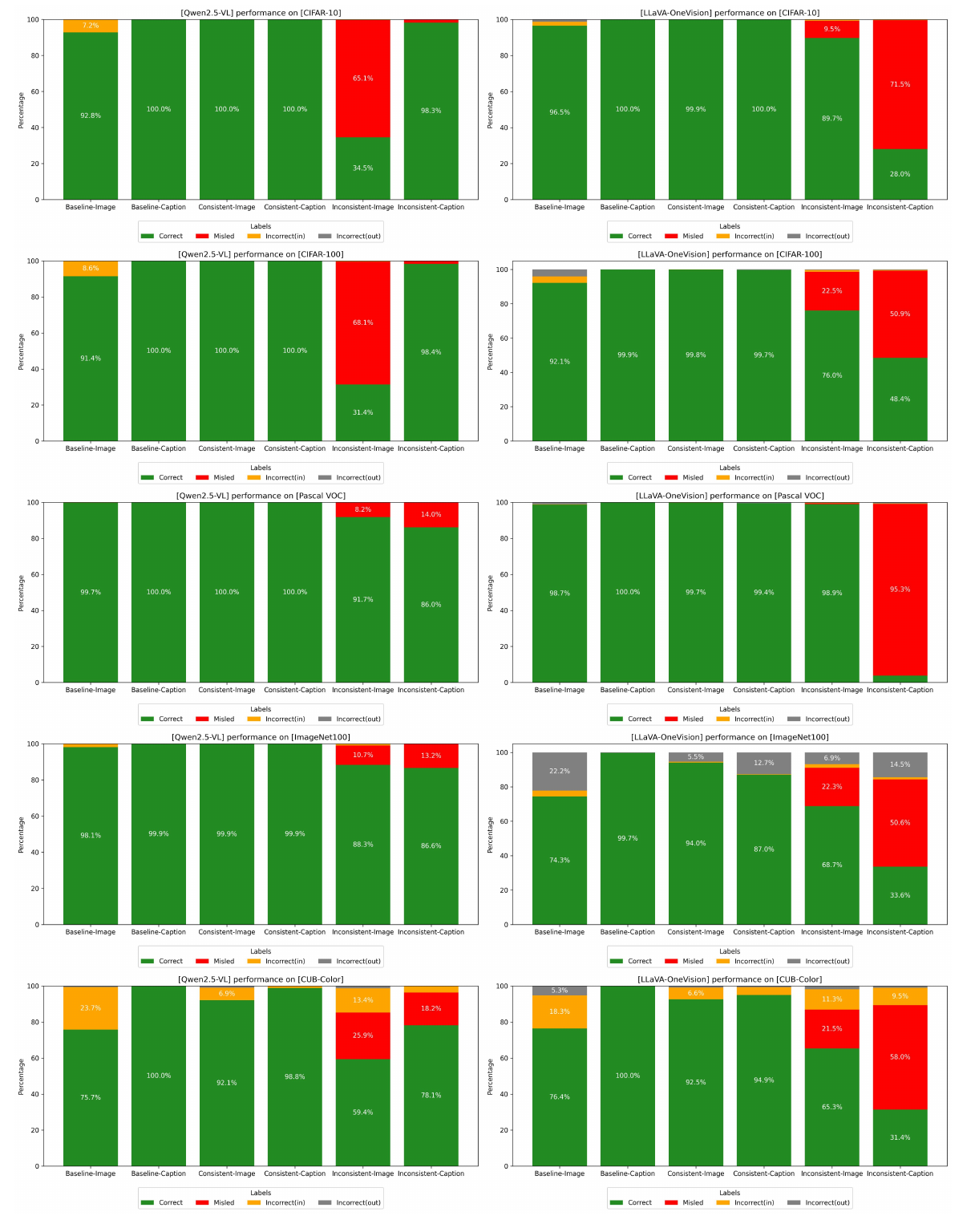}
    \caption{Breakdown of model predictions under unimodal baselines, consistent input pairs and inconsistent input pairs. Left: Qwen2.5-VL; right: LLaVA-OneVision.}
    \label{fig:breakdown_qwen_ov}
\end{figure}

Previously in Figure~\ref{fig:behavioral_results} and Figure~\ref{fig:behavior_consistent}, we presented the accuracy of the predictions under inconsistent and consistent input pairs in comparison to the single-modality input baselines. Here, we further showcase the breakdown of the predictions into 1) correct classes, 2) misled classes (which match the class labels of the inputs from the non-target modality), 3) in-option incorrect class labels (any class labels given in the options but does not belong to the image and caption class labels) and 4) out-of-option predictions. We include the results for every model and dataset pair in Figure~\ref{fig:breakdown_llava_blip} and Figure~\ref{fig:breakdown_qwen_ov}. 

\clearpage

\section{Additional Representational Analysis Results}
\label{app:additional_representational_analysis_results}
We run experiments on all models on all datasets that we have considered. Below is a list of figures/tables for the corresponding experiment.

\subsection{Hypothesis 1: Does the model fail to encode the individual modalities?}
\begin{figure}[p]
    \centering
    \includegraphics[width=1.0\linewidth]{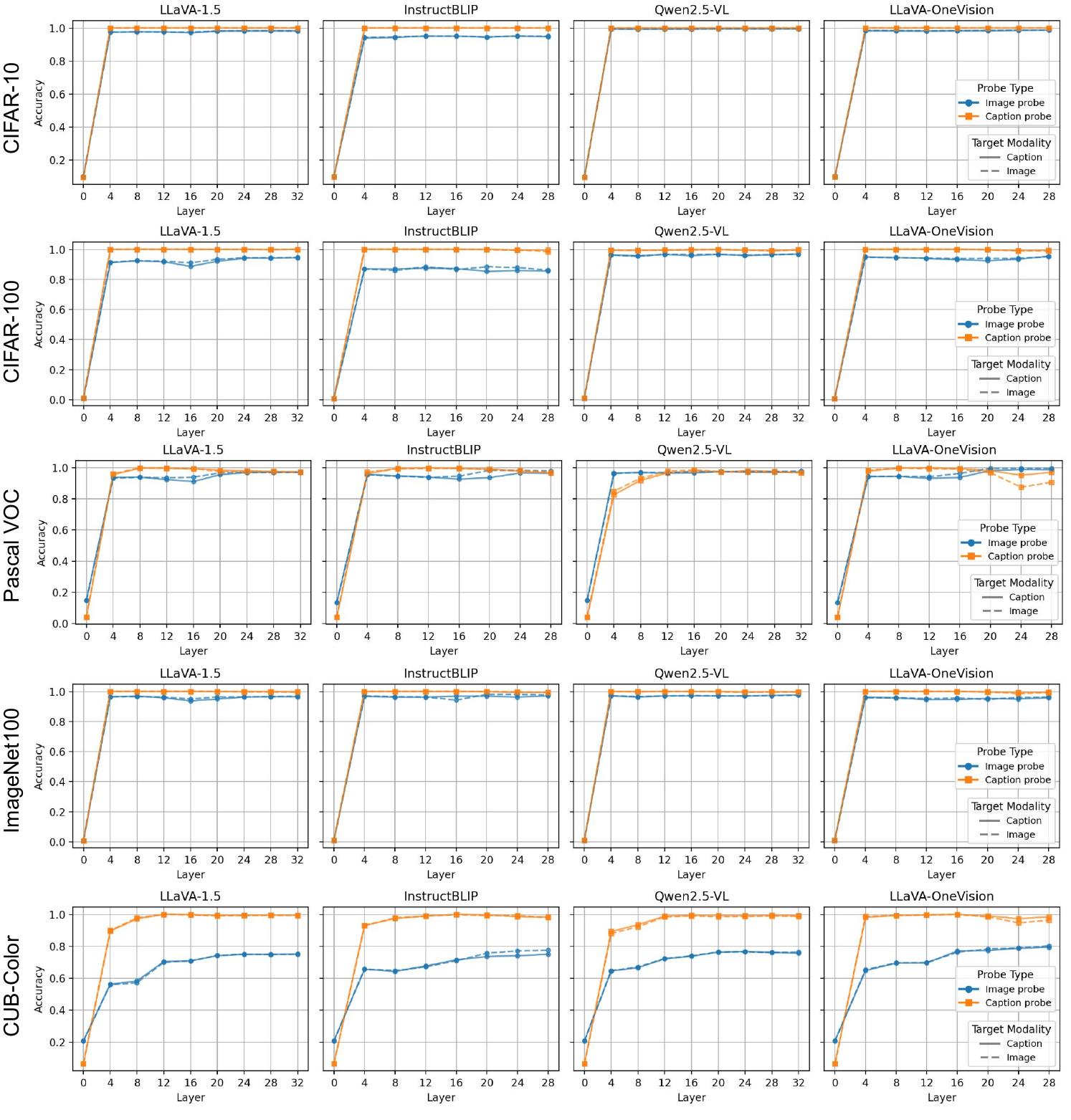}
    \caption{Probe accuracies across layers for all model-dataset pairs, trained on inconsistent inputs to extract unimodal information.}
    \label{fig:unimodal_probe_all}
\end{figure}

See full results in Figure \ref{fig:unimodal_probe_all}.

\subsection{Hypothesis 2: Does the model fail to detect that the modalities are in conflict with one another?}
\begin{figure}[p]
    \centering
    \includegraphics[width=1.0\linewidth]{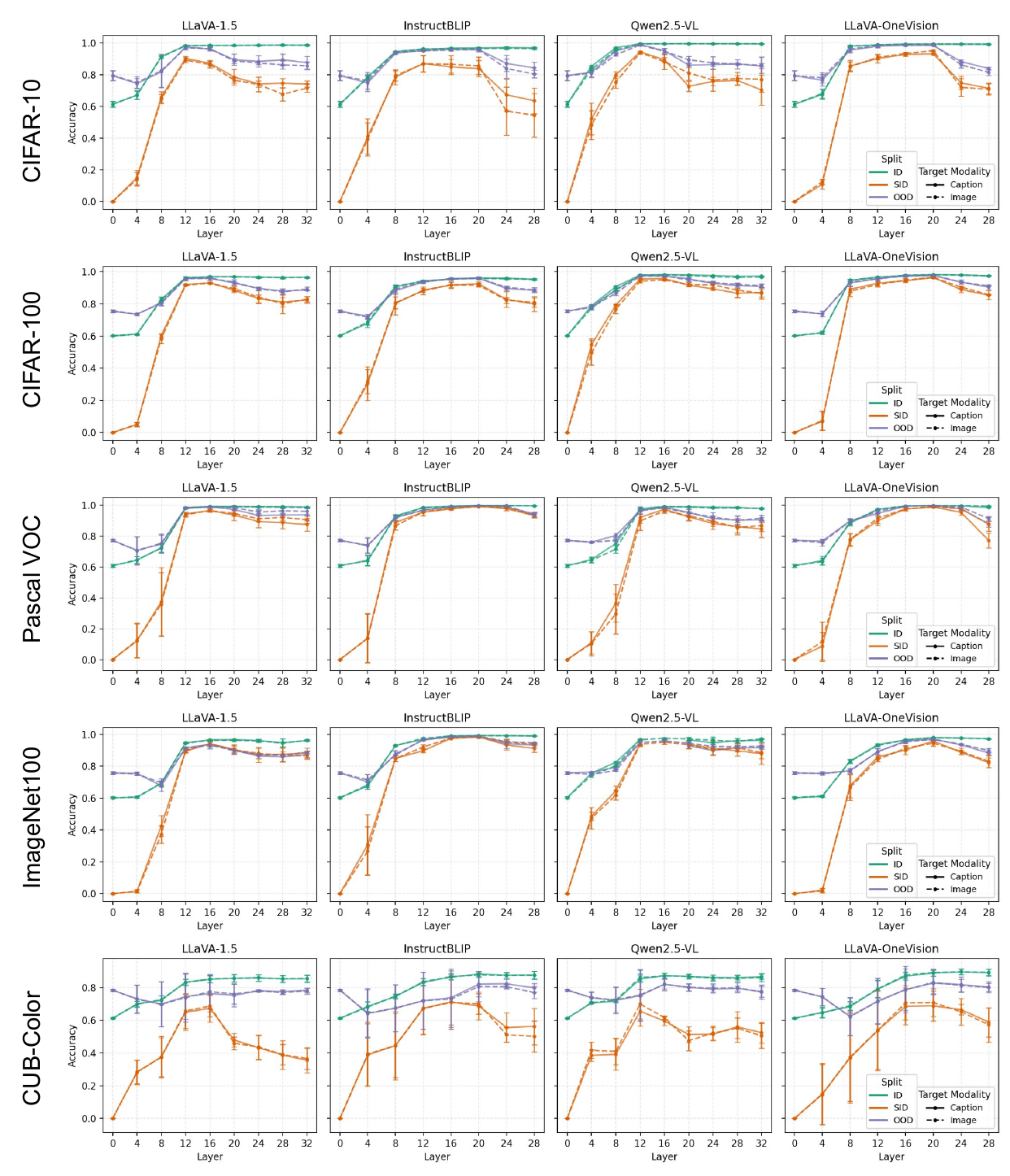}
    \caption{Consistency probe accuracies across layers for all model-dataset pairs.}
    \label{fig:consistency_probe_all_pairs}
\end{figure}

See full results in Figure \ref{fig:consistency_probe_all_pairs}.

Section~\ref{sec:alignment_probe} presents the consistency probe accuracy. Here, we want to study whether the trained probes accurately capture the idea of consistency rather than overfitting to the distribution of training data. Here we perform an additional 3-fold evaluation over the class label space (on top of data-wise the train-test split)  -- we divide the class labels into three disjoint subsets. For each fold, the consistency probe is trained on image-caption combinations whose labels come from two of the subsets. At test time, the probe is evaluated on both (1) image-caption pairs with labels drawn from the same label subsets used in training (in-distribution data, hereafter ID), (2) image-caption pairs composed from labels in the held-out subset not seen during training (out-of-distribution data, hereafter OOD), and (3) image-caption pairs with exactly one label in the training class label subsets paired with one label in the held-out class label subset (semi-in-distribution, hereafter SID). This setup allows us to evaluate whether the probe's learned consistency signal generalizes beyond specific class-label combinations to encode more general notions of consistency across input modalities. 

Figure~\ref{fig:consistency_probe_all_pairs} shows the consistency probe accuracy for all model-dataset pairs. We observe a similar trend across models and across datasets: the ID accuracy increases across layers and achieves a high accuracy (nearly 1.0 for all datasets except CUB-Color), the OOD accuracy is initially higher than the ID accuracy for the very early layer and then overlaps with the ID accuracy in the middle layers before it suffers a small drop towards the final layers. Comparing the ID and OOD accuracies, it suggests that the probe is able to generalize and learn the concept of consistency for unseen class-combinations, especially during the middle-layer representations. We also noticed that the SID accuracy is very low (starting with 0\%) in the early layers and we suspect that this is due to it containing only the inconsistent pairs thus has a very different distribution as the training ID data distribution, which makes the predictions systematically biased towards predicting them to be consistent.

\subsection{Hypothesis 3: Does the model inherently favor one modality over the other in representational space?}
\begin{figure}[p]
    \centering
    \includegraphics[width=1.0\linewidth]{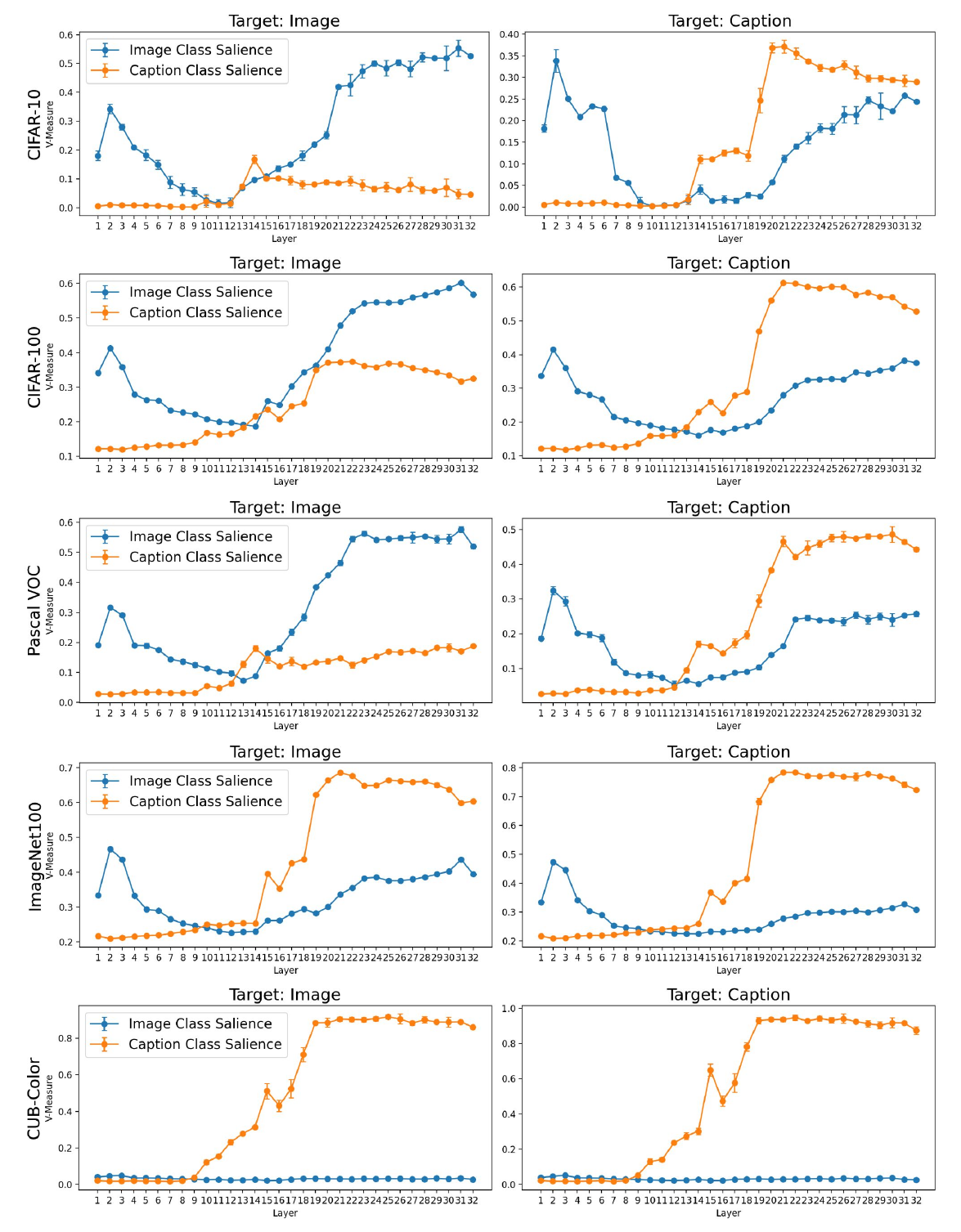}
    \caption{Layer-wise V-Measure of LLaVA-1.5 representations with regard to the image and caption labels of the inconsistent samples, on all datasets. V-Measure is computed over K-Means clustering models initialized with three random seeds.}
    \label{fig:all_vmeasure_diff_llava1.5}
\end{figure}

\begin{figure}[p]
    \centering
    \includegraphics[width=1.0\linewidth]{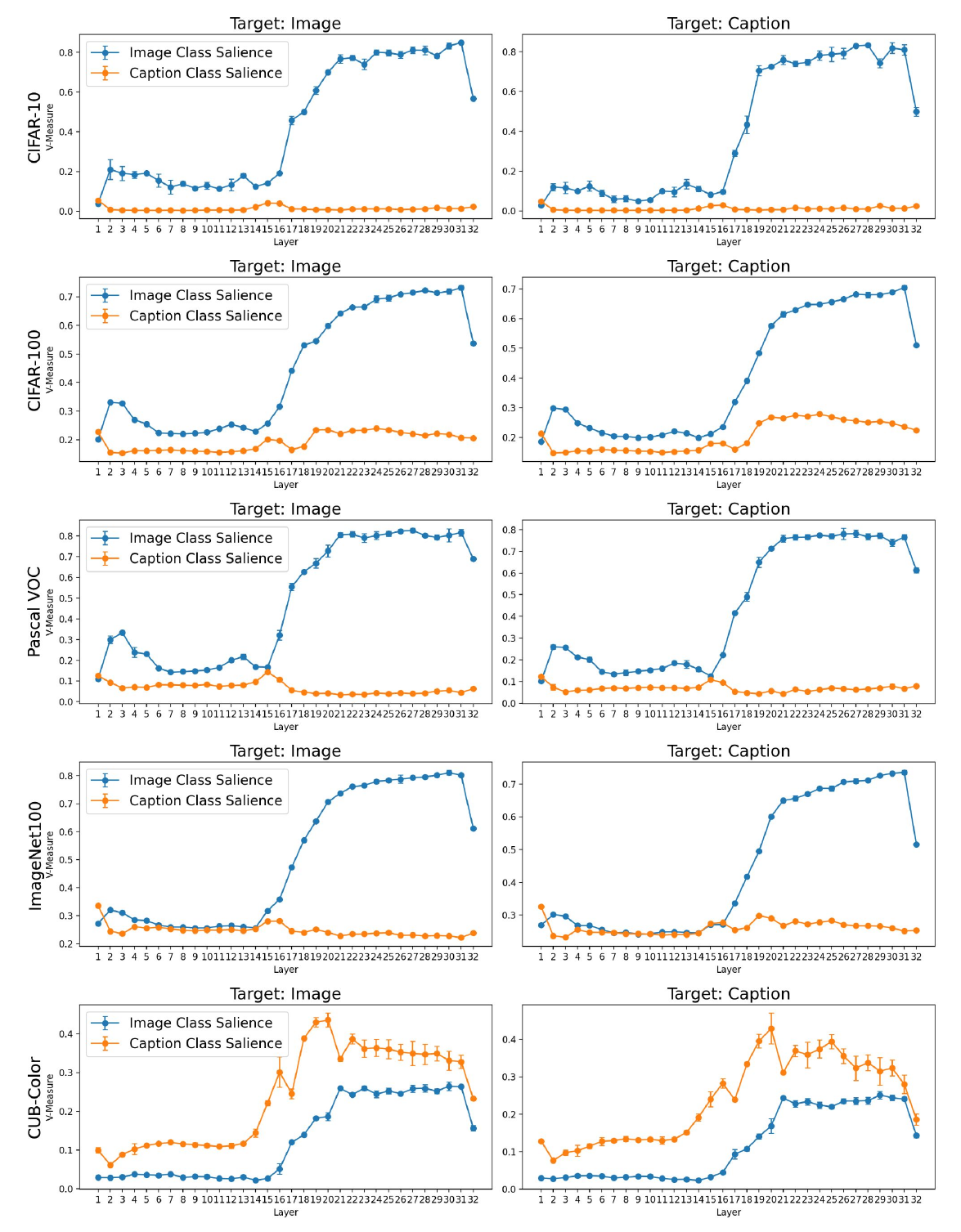}
    \caption{Layer-wise V-Measure of InstructBLIP representations with regard to the image and caption labels of the inconsistent samples, on all datasets. V-Measure is computed over K-Means clustering models initialized with three random seeds.}
    \label{fig:all_vmeasure_diff_blip}
\end{figure}

\begin{figure}[p]
    \centering
    \includegraphics[width=1.0\linewidth]{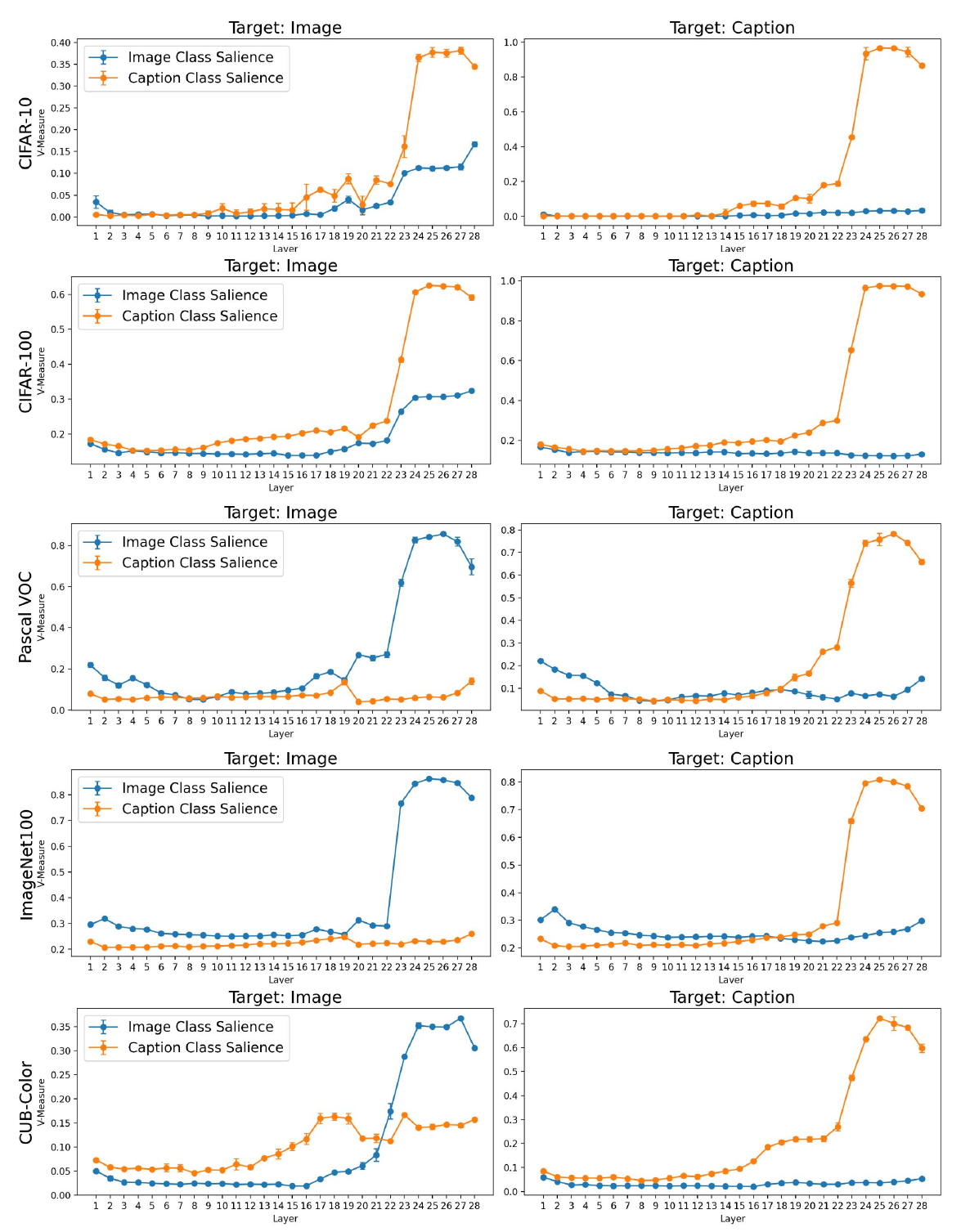}
    \caption{Layer-wise V-Measure of Qwen2.5-VL representations with regard to the image and caption labels of the inconsistent samples, on all datasets. V-Measure is computed over K-Means clustering models initialized with three random seeds.}
    \label{fig:all_vmeasure_diff_qwen}
\end{figure}

\begin{figure}[p]
    \centering
    \includegraphics[width=1.0\linewidth]{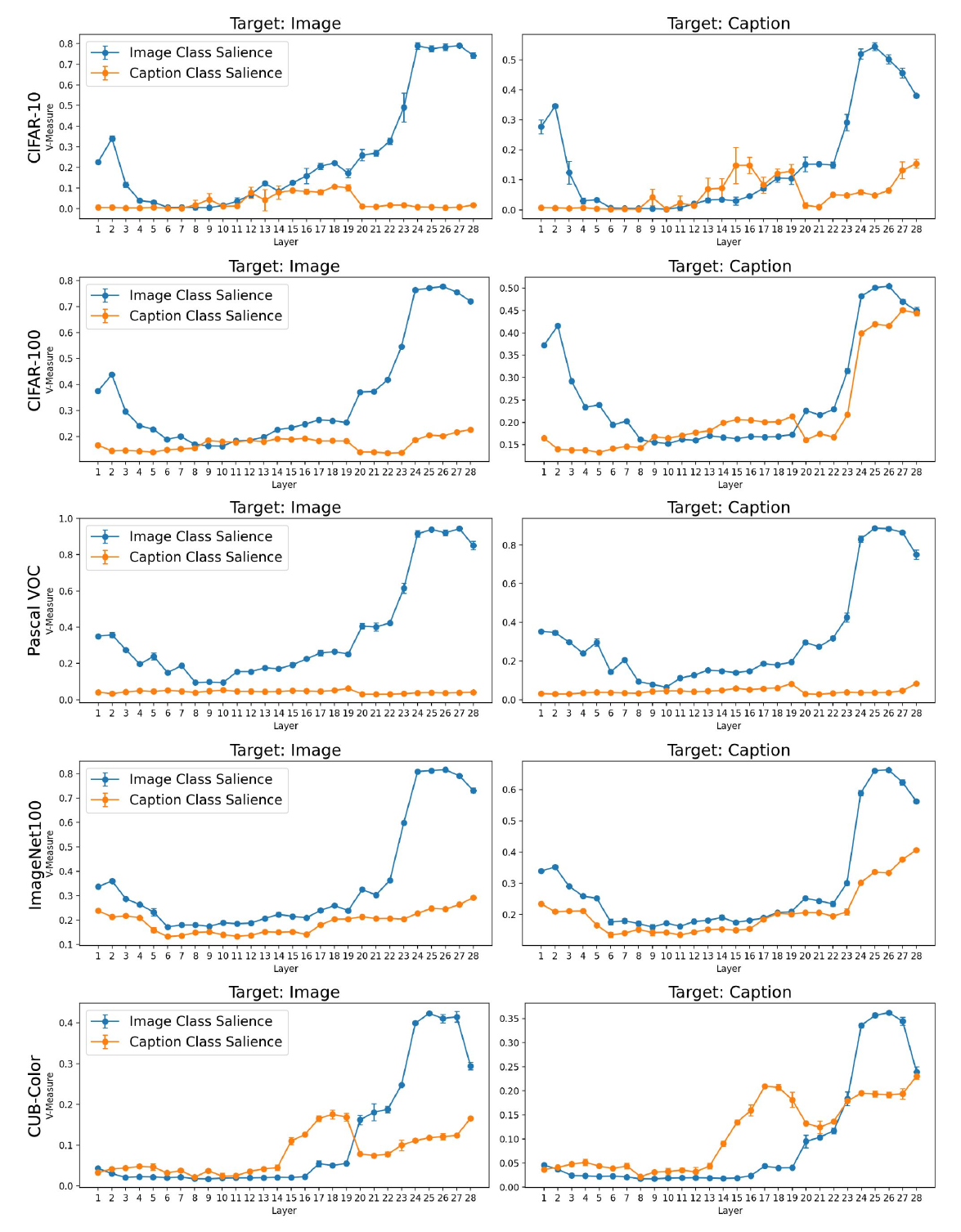}
    \caption{Layer-wise V-Measure of LLaVA-OneVision representations with regard to the image and caption labels of the inconsistent samples, on all datasets. V-Measure is computed over K-Means clustering models initialized with three random seeds.}
    \label{fig:all_vmeasure_diff_llava_ov}
\end{figure}

See full results in Figures \ref{fig:all_vmeasure_diff_llava1.5}, \ref{fig:all_vmeasure_diff_blip}, \ref{fig:all_vmeasure_diff_qwen}, and \ref{fig:all_vmeasure_diff_llava_ov}.

\clearpage

\section{Additional Attention Head Intervention Results}
\label{app:additional_attention_head_intervention_results}

\subsection{Classification of Attention Heads}
\label{app:classification_of_attention_heads}

We use the following criteria to identify whether an attention head is a router head, an image promotion head or a caption promotion head:

We compute the change in performance over 21 $\alpha$ values linearly ranging from -10 to 10,  recording both the portion of predictions that correspond to the target modality class label as well as the ones that correspond to the non-target modality class label. Then, we look at the respective performance changes when the target modality is image and when the target modality is caption. For modality-agnostic router heads, target modality prediction should both increase as $\alpha$ increases. For image promotion heads, target modality prediction should increase as $\alpha$ increases when the target modality is image and should drop when the target modality is caption. For caption promotion heads, target modality prediction should increase as $\alpha$ increases when the target modality is caption and should drop when the target modality is image.

The increasing trend can be implemented by a \texttt{monotonically\_increasing} function, with some epsilon allowing small, local drops as long as the trend is increasing. The decreasing trend can be implemented by a \texttt{monotonically\_decreasing} function, with some epsilon allowing small, local rises as long as the trend is decreasing. Apart from these two essential filters, we also make sure that any drop in reporting answers of one modality correspond to a rise in the other modality, so that the intervention of an attention head only changes the \textit{preference} of which modality that Qwen2.5-VL reports but not degrade its overall ability to generate well-formed English utterances.

We noticed although these stated criteria are intuitive, the actual intervened accuracies across $\alpha$ for each model can be noisy and involve non-monotonic trends. Because of this, we report only the attention heads exhibiting the most salient trends or the most intervenability as weaker signals might be misclassified into multiple head types and will be less interesting to further analyze. 


\subsection{Discussion on Classification and Intervenability of Attention Heads}
\label{app:discussion_of_attention_heads}

\begin{figure}
    \centering
    \includegraphics[width=1.0\linewidth]{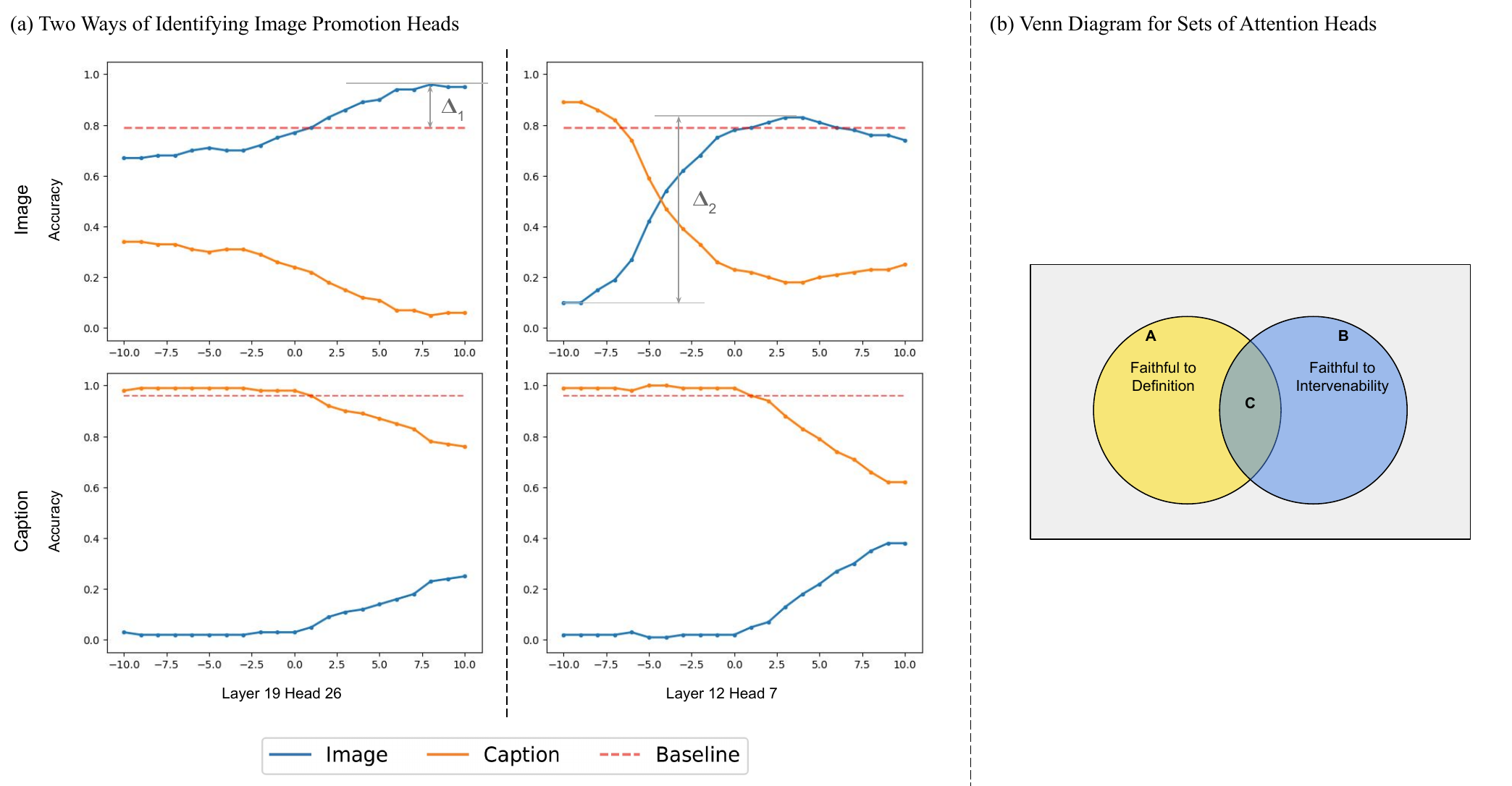}
    \caption{Fine-grained classification of attention heads. (a) We show two ways to rank the importance heads (i.e., we use image promotion heads for illustration purpose): (1) $\Delta_1$ is the \textit{intervenability}, which is the performance improvement a model can have after the intervention; (2) $\Delta_2$ is the criteria we use to define the identity of attention heads in Section~\ref{sec:head_types}. (b) A Venn diagram that shows different sets of attention heads. Set \textbf{A} represents the heads that follow our definition ($\Delta_2$), while set \textbf{B} represents those that prioritize intervenability. It is important to note that these two sets of heads do not necessarily completely overlap. In the ideal picture, we would like to find the components of a model (i.e., attention heads in our work) that fall in the overlapping set \textbf{C}, which satisfies both criteria we have set.}
    \label{fig:app_head_discussion}
\end{figure}

Apart from filtering, we also explore two ways of ranking the importance of heads, one focuses on the intervenability of the heads -- if a head can be intervened to improve unimodal or cross-modal performance beyond the unintervened baseline, we consider that head to be ``intervenable''; the other one focuses how salient are the aforementioned trends exhibited in the accuracy plots -- if across $\alpha$ values a head brings the most significant change in accuracy, we consider it to most saliently exhibit its head type's traits. Figure~\ref{fig:app_head_discussion} left shows a side by side comparison between two image promotion heads: L19H26 is considered to be highly intervenable, with a large $\Delta_1$ compared to the image baseline while L12H7 shows a very salient pattern (indicated by the large $\Delta_2$) consistent with our expectation of the trait of an image promotion head. 

It is important to note that these two types of attention heads do not necessarily overlap completely with one another. Figure~\ref{fig:app_head_discussion} demonstrates different sets of attention heads. The set \textbf{A} refers to the image promotion head L12H7 which faithfully follows the definition from Section~\ref{app:discussion_of_attention_heads}, while the set \textbf{B} refers to the image promotion head L19H26 that prioritize the intervenability. In the ideal picture, we would like to find the components of a model (i.e., attention heads in our work) that fall in the overlapping set \textbf{C}, which satisfies both criteria we have set. L19H26 belongs to set \textbf{C}, indicating that (1) across $\alpha$ values, it can correspondingly impact models' behaviors (i.e., the larger $\alpha$ is, the more frequently image information is reported), and (2) intervening this head can bring significant performance improvement.

\subsection{Distribution of Modality-Agnostic Router Heads \& Modality-Specific Promotion Heads}
\label{app:modality_specific_promotion_heads}

\begin{figure}
    \centering
    \includegraphics[width=\linewidth]{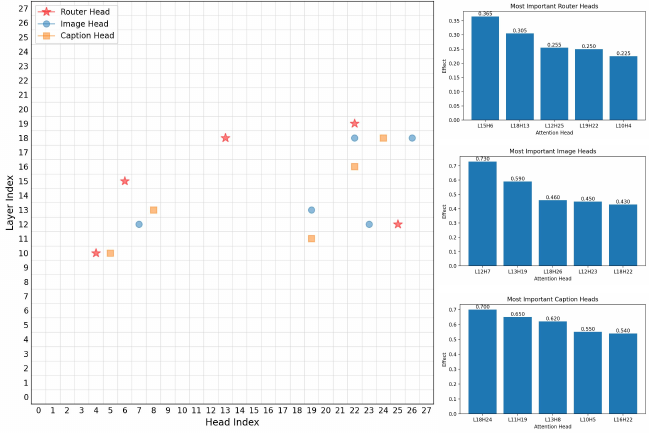}
    \caption{Top-5 heads of each type showing best trends consistent with head traits. Left: the layer-wise location of each of the head; right: the maximum unimodal or cross-modal change in accuracy that each head exhibits upon intervention.}
    \label{fig:Good_trend_heads}
\end{figure}

\begin{figure}
    \centering
    \includegraphics[width=\linewidth]{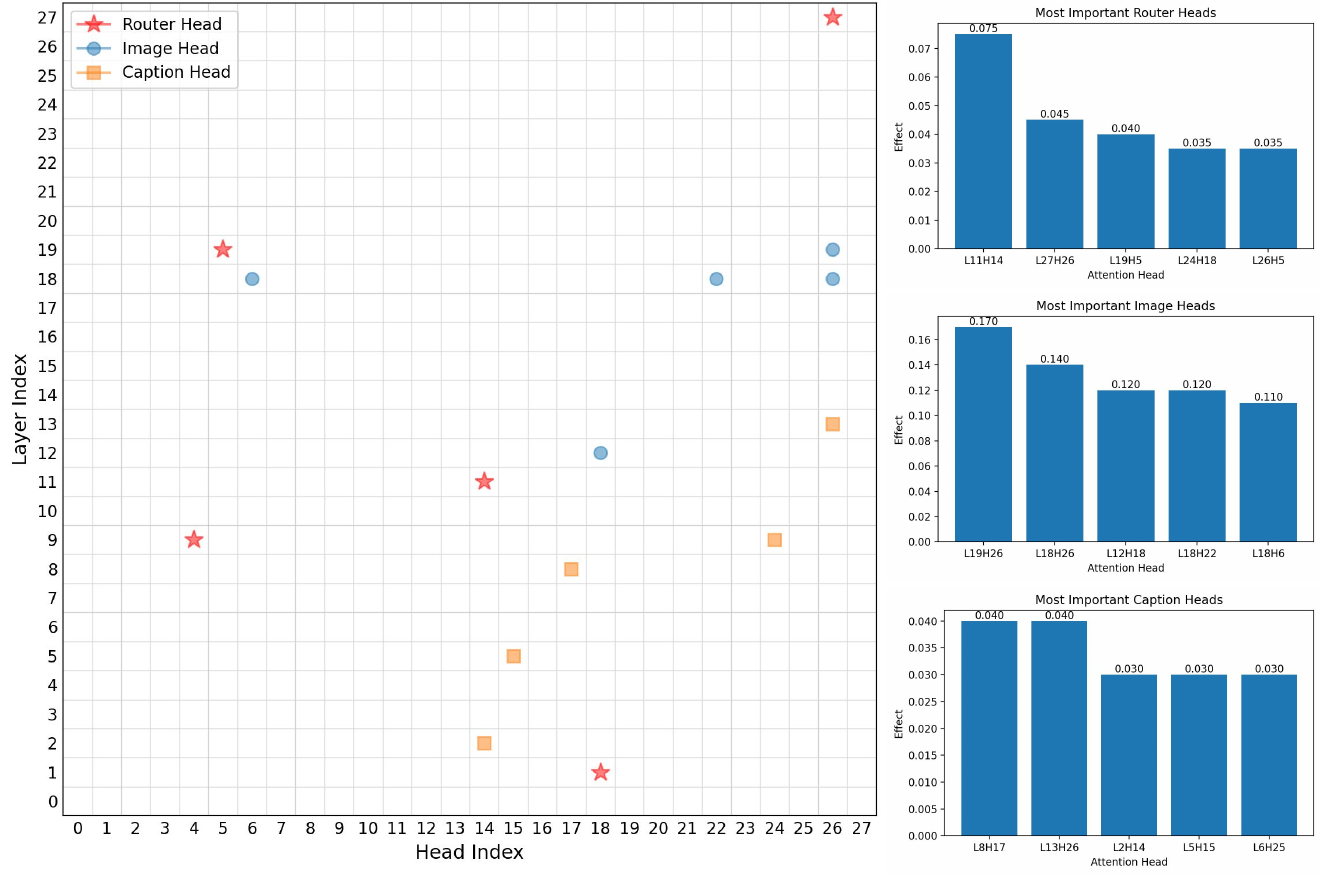}
    \caption{Top-5 heads of each type showing most intervenability. Left: the layer-wise location of each of the head; right: the maximum unimodal or cross-modal improvements that each head exhibits upon intervention, compared with the unintervened baselines.}
    \label{fig:Intervenable_heads}
\end{figure}

Here, we present the top-5 most important heads of each type based on the trend salience (Figure~\ref{fig:Good_trend_heads}) and intervenability (Figure~\ref{fig:Intervenable_heads}).

\subsection{Additional Post-Intervention Representational Salience Plots}
\label{app:representational_salience_plots}

\begin{figure}[p]
    \centering
    \includegraphics[width=\linewidth]{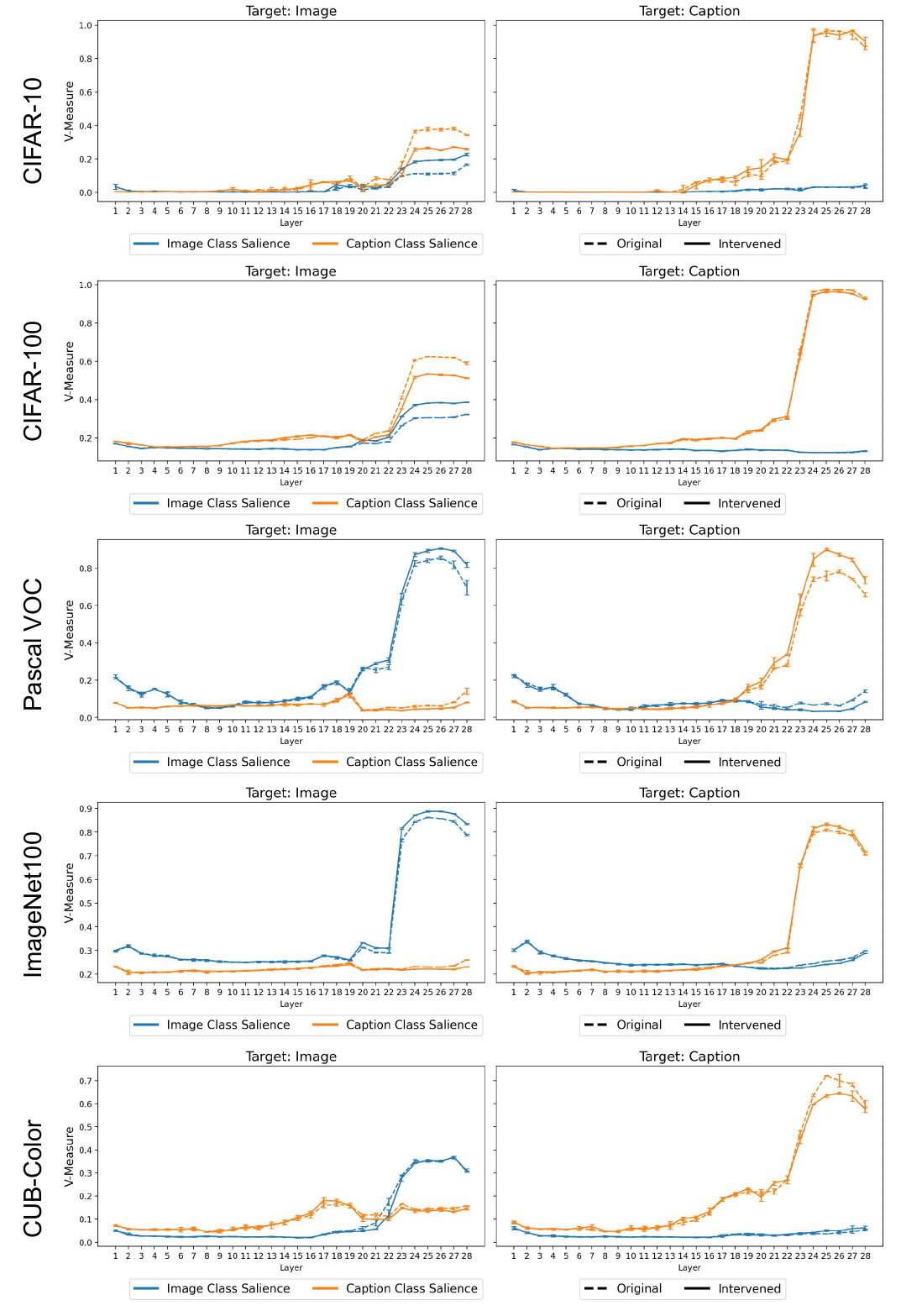}
    \caption{Representational salience of image and caption classes before and after intervening the router head L11H14 of Qwen2.5-VL.}
    \label{fig:router_cluster_all}
\end{figure}

\begin{figure}[p]
    \centering
    \includegraphics[width=\linewidth]{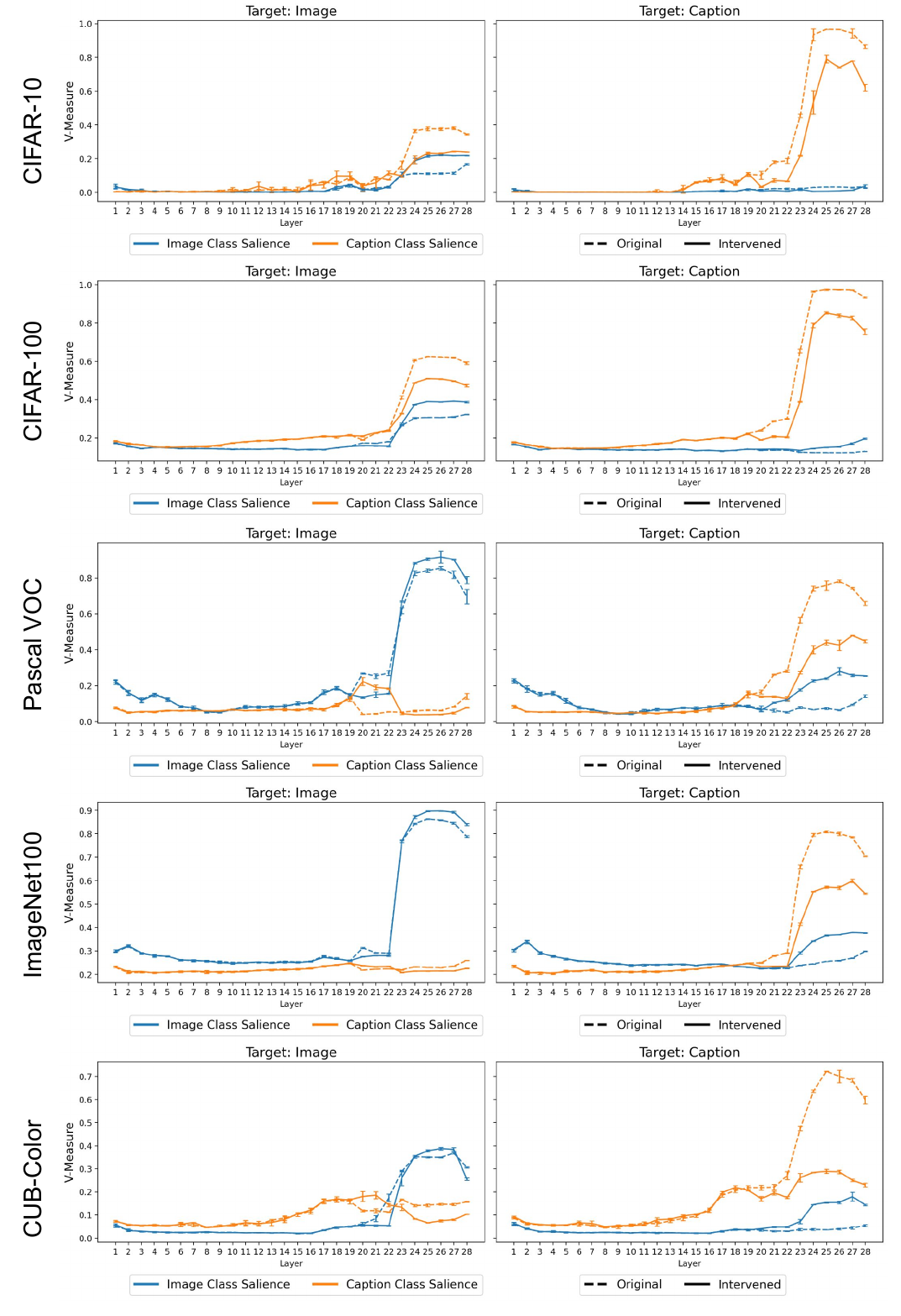}
    \caption{Representational salience of image and caption classes before and after intervening the image promotion head L19H26 of Qwen2.5-VL.}
    \label{fig:image_pro_cluster_all}
\end{figure}

\begin{figure}[p]
    \centering
    \includegraphics[width=\linewidth]{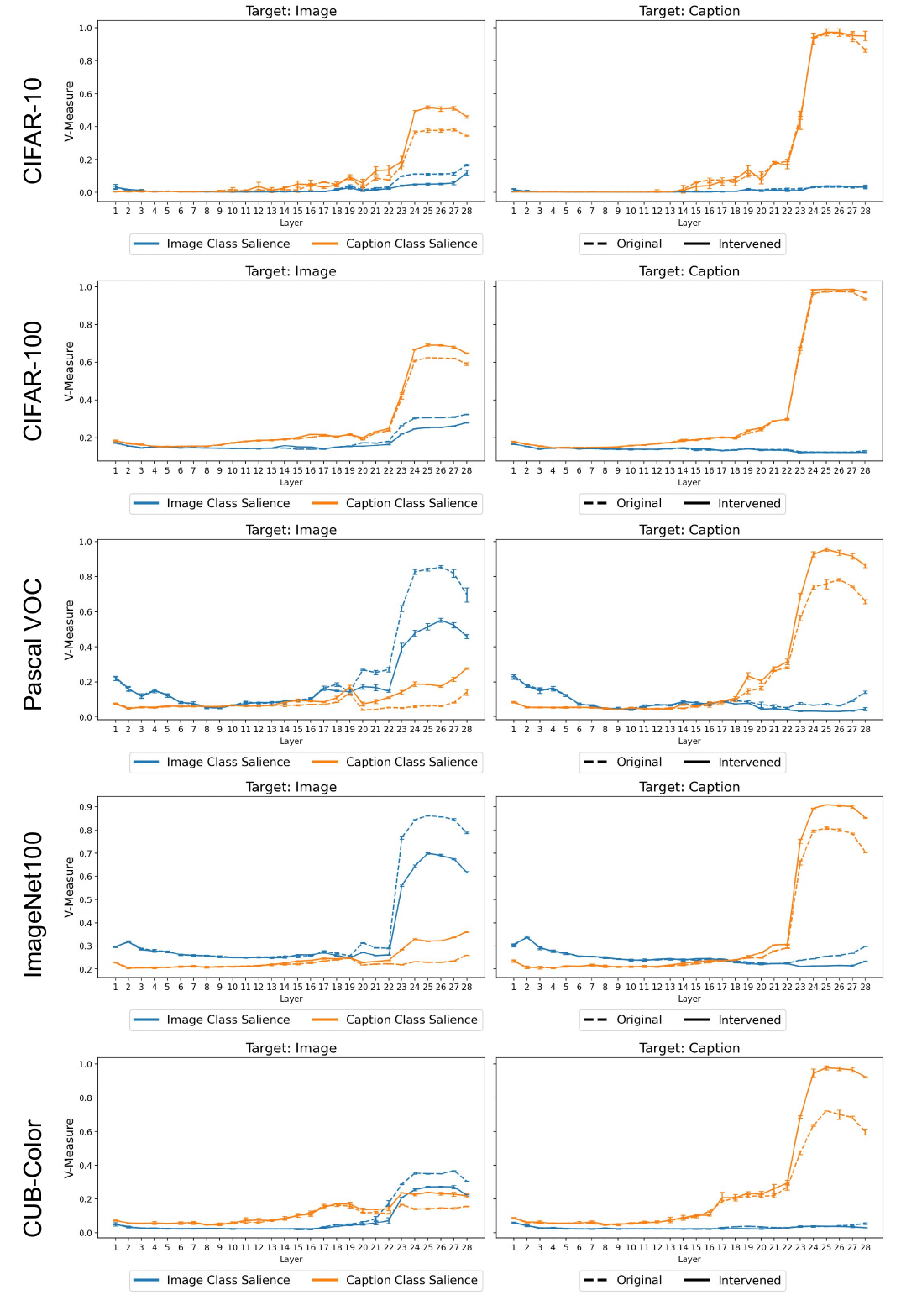}
    \caption{Representational salience of image and caption classes before and after intervening the caption promotion head L13H26 of Qwen2.5-VL.}
    \label{fig:caption_pro_cluster_all}
\end{figure}

In Figure~\ref{fig:router_cluster_all}, Figure~\ref{fig:image_pro_cluster_all} and Figure~\ref{fig:caption_pro_cluster_all}, we present the effect of intervening ($\alpha=10$) the router head L11H14, image promotion head L19H26, and caption promotion head L13H26 of Qwen2.5-VL, on all datasets and target modalities.

\end{document}